# Dynamic Multivariate Functional Data Modeling via Sparse Subspace Learning

Chen Zhang[1], Hao Yan[2], Seungho Lee[3], Jianjun Shi[4]

[1]National University of Singapore, Singapore, 117576

[2]Arizona State University, Tempe, AZ, 85282

[3]Samsung Electronics, Suwon, South Korea, 443742

[4]Georgia Institute of Technology, Atlanta, GA, 30332

## Abstract

Multivariate functional data from a complex system are naturally high-dimensional and have complex cross-correlation structure. The complexity of data structure can be observed as that (1) some functions are strongly correlated with similar features, while some others may have almost no cross-correlations with quite diverse features; and (2) the cross-correlation structure may also change over time due to the system evolution. With this regard, this paper presents a dynamic subspace learning method for multivariate functional data modeling. In particular, we consider different functions come from different subspaces, and only functions of the same subspace have cross-correlations with each other. The subspaces can be automatically formulated and learned by reformatting the problem as a sparse regression. By allowing but regularizing the regression change over time, we can describe the cross-correlation dynamics. The model can be efficiently estimated by the fast iterative shrinkage-thresholding algorithm (FISTA), and the features of every subspace can be extracted using the smooth multi-channel functional PCA. Numerical studies together with case studies demonstrate the efficiency and applicability of the proposed methodology.

# 1    Introduction

Multivariate functional data, which arise from a collection of simultaneous recordings of several time courses for many subjects or units, are increasingly common and important in various applications. One concrete motivating example we introduce here is human gesture tracking (Lui 2012; Wang et al. 2012). In particular, the movements of a human subject are tracked by capturing real-time positions of 18 body joints using a Kinect pose estimation pipeline. Its data acquisition rate is 30Hz with 2cm accuracy in joint positions. Every joint is recorded as a point in a three-dimensional Cartesian coordinate system. Treating every coordinate of every joint as one function, we totally have $18 \times 3$ functions. The subject is instructed to conduct two gestures "bow up" and "throw" sequentially with total 248 frames captured. Figure 1 shows eight selected frames during these two gestures. The values of these 54 functions observed at these 248 time points are also shown in Figure 2. Clearly, some functions share very similar features with each other (such as Functions 4, 7, and 10 in Figure 3), indicating that they are strongly correlated. This is because these joints move in similar ways, such as the six joints on the two arms. In contrast, some other functions have quite diverse features with each other (such as Functions 30, 47 and 51 in Figure 4), indicating that they are not correlated with each other. This is because that these joints move in different ways, such as one joint on the arm and another joint on the leg. With this regard, we may infer that these functions can be naturally clustered into different groups. Equivalently, we can say that these functions lie in different subspaces. Another thing to be noted is that their cross-correlation structure changes over time. For example, Functions 18 and 33 (shown in Figure 5a) only share similar patterns in the first 110 time points. Then their cross-correlation disappears. On the contrary, Functions 4 and 13 (shown in Figure 5b) have quite diverse patterns in the first 110 time points, but then they begin to move similarly in the subsequent 138 time points. This is because that the first 110 time points are from the "bow up" gesture, while the last 138 points are from the "throw" gesture. Since for different gestures, the joints are required to cooperate and move in different ways, their cross-correlation structure would change.

In addition to the example shown above, multivariate functional data exist in many other applications. For example, in traffic monitoring, many traffic variables such as vehicle speed, flow rate, occupancy, *etc*, are continuously recorded for anomaly detection. In semiconductor manufacturing systems, hundreds of sensors are installed in a chamber to real-time monitor different process variables (e.g. temperature, pressure, electronic flows, *etc*). In electroencephalography (EEG) tests, multiple electrodes are placed at different places to record the brain activity over a certain time period for epilepsy signal detection. Therefore, there is a pressing need to model and analyze those multichannel data with consideration of their data complex cross-correlation structure and dynamics.

The remainder of the paper is organized as follows. Section 2 reviews the state-of-the-art methods on multivariate functional data analysis. Section 3 introduces our proposed dynamical functional subspace model in detail. Section 4 discusses the model inference procedure. Section 5 uses some numerical studies to demonstrate the advantages of the proposed model by comparing it with some other state-of-the-art methods. Section 6 applies the developed methods into two real-data examples from human gesture tracking experiments and manufacturing systems. Finally, Section 7 concludes this paper with remarks. Some technical details are provided in the Appendices.

## 2    Literature Review

This section will review the methodology on the correlation analysis of multivariate functional data. In the literature, many existing works for multivariate functional data analysis focus on bivariate functions, by developing measures to quantify their cross-correlations. In particular, Hannan (1961) introduced the canonical correlation analysis to measure the distance between two functions based on their subspace angles. Later, Leurgans et al. (1993) revised it by considering the smoothness of canonical variates. Eubank and Hsing (2008) proposed to extend the canonical analysis from the original functional space to the reproducing kernel

Hilbert space for more comprehensive feature comparison. Recently, Dubin and Müller (2005) proposed the dynamical correlation analysis based on the scalar product of functions after normalization. Yang et al. (2011) considered functional singular value decomposition and used the singular functions to measure cross-correlations.

As to multivariate functional data analysis (e.g. for more than two functions), most existing works apply dimension reduction analysis for function feature extraction and depiction through various approaches. In particular, for linear cross-correlations, Pan and Yao (2008) proposed a dynamic factor model to extract common factors from multiple functional data. Di et al. (2009); Chiou et al. (2014); Paynabar et al. (2016) introduced several multi-channel functional principal component analyses (MFPCA) methods to describe the within-function and between-function correlations. Chiou and Müller (2016) proposed a pairwise interaction model based on the cross-covariance surfaces between functions. For nonlinear cross-correlations, Chiou and Müller (2014) used a functional manifold model to regularize the functional features and characterize the cross-correlations. Besides dimension reduction techniques, other works include adapting the traditional parametric random-effects models (Fieuws and Verbeke 2006), and the nonparametric kernel smoothing approach for individual function modeling (Xiang et al. 2013). However, one common limitation of all the aforementioned methods is that they assume multivariate functions are strongly correlated since each function is assumed to be a linear combination of all the extracted features. This assumption leads these methods to fail to recover correct functional features and fail to model the multivariate functions accurately when they have diverse features with sparse cross-correlations, i.e., come from different clusters.

As a more powerful tool to describe conditional dependence structures between different random variables, graphical models are recently used for multivariate functional data analysis to represent their sparse partial cross-correlations. In particular, Qiao et al. (2015) extended the graphical LASSO (Yuan and Lin 2007) to multivariate functional data by constructing a penalized log-Gaussian likelihood on the functional PCA scores. Zhu et al. (2016) proposed a Bayesian framework by first applying the functional PCA and then applying the Markov distributions and hyper Markov laws on the extracted PCA scores for graphic decomposition. However, all the methods based on graphical LASSO need to assume that the functions are Gaussian processes, which may not be true in practice. In literature, another type of methods to represent the sparse partial cross-correlations of multivariate functions is based on sparse subspace clustering. For example, Bahadori et al. (2015) applied the sparse subspace clustering method (Elhamifar and Vidal 2013) into functional data analysis. This method directly uses the self-expression representation to describe the partial cross-correlations of multivariate functional data and regularizes the sparsity of the regression coefficients. However, all the aforementioned methods cannot tackle functions with dynamic cross-correlations.

Considering that functional data in most real applications are sampled over a grid of time points, one possible approach to handle this sort of dynamic cross-correlations is to construct separate graphs for every time point, and regularize these graphs to be temporal consistent. Specifically, Zhou et al. (2010); Kolar and Xing (2011); Qiu et al. (2016) proposed to use a nonparametric kernel smoothing approach. However, the model estimation for every time point with kernel smoothing requires extremely heavy computation, which hinders its application in cases with large sample size or high dimensions. Kolar and Xing (2012) proposed to use $l_2$ penalty to regularize change of the graphs. However, this model still assumes every variable (dimension) is Gaussian distributed. Furthermore, these two models still give too much flexibility for the graph dynamics. Even for some time intervals where no cross-correlation changes at all, these models still generate fictitious dynamics, consequently leading to the biased estimation of the cross-correlation change points.

Motivated by the wide applications of multivariate or even high-dimensional functional data with sparse and dynamic cross-correlations, and the infancy of reasonable models to describe them, this paper further explores this field with twofold contributions. First, we propose a dynamic model for multivariate functional data based on sparse subspace learning. In particular, we consider different functions come from different subspaces in the sense that only functions of the same subspace have nonzero cross-correlations with each other. This subspace relationship can be learned as a sparse self-expressive linear regression. Second, to describe the cross-correlation dynamics, we allow the regression coefficients to change over time, but regularize the change variability with a fused LASSO penalty. This model can be efficiently estimated by the fast iterative shrinkage-thresholding algorithm (FISTA) with the fused LASSO solver. Based on the model, we can cluster the functions into different subspaces, compute the change point of the functions and extract the subspace features using a proposed smooth multi-channel functional PCA. Numerical studies together with two real case studies demonstrate the efficiency and applicability of the proposed model.

# 3 Dynamic Functional Subspace Learning

In this section, we will first review the work on sparse subspace clustering and how it can be used to model the sparse cross-correlation functions in Section 3.1. We then extend this framework to the dynamic correlation structure with the proposed dynamic functional subspace learning via the fused Lasso penalty in Section 3.2. Its theoritical properties are discussed in Section 3.3.

## 3.1 Modeling Multivariate Functions via Sparse Subspace Learning

Consider a $p$-dimensional (e.g. $p$ channels) functional sample with the $i^{th}$ sample $\mathbf{Y}_i(t) = [Y_{i1}(t), \ldots, Y_{ip}(t)]$ whose $t$ is on a compact interval $\mathcal{T} = [0, T]$, such that $\int_{\mathcal{T}} \mathrm{E}\left[Y_{ij}(t)^2\right] \mathrm{d}t < \infty$ for $j = 1, \ldots, p$. In particular, we assume

$$Y_{ij}(t) = X_{ij}(t) + \epsilon_{ij}(t), \tag{1}$$

where $X_{ij}(t)$ is the signal function and $\epsilon_{ij}(t)$ is the independent noise function with mean $\mathrm{E}(\epsilon_{ij}(t)) = 0$ and bounded signal-to-noise ratio $\sigma^2 = \int_{\mathcal{T}} \epsilon_{ij}(t)^2 \mathrm{d}t / \int_{\mathcal{T}} X_{ij}(t)^2 \mathrm{d}t$. The noise can also have autocorrelation $\Gamma_j(t, s) = \mathrm{E}(\epsilon_j(t), \epsilon_j(s))/\sigma^2$. Furthermore, we assume that these $p$ signal functions $\mathbf{X}_i(t) = [X_{i1}(t), \ldots, X_{ip}(t)]$ can be partitioned into $L$ different subspaces, $\mathcal{S}_l, l = 1, \ldots, L$. The functions in the same subspace have strong cross-correlations, while the functions in different subspaces have no cross-correlations.

**Assumption 1.** *((A1) Subspace Assumption)*

*Each subspace $\mathcal{S}_l$ is defined as the set of all functions linearly combined by $d_l$ basis functions $\mathbf{\Phi}_l = [\phi_{l1}(t), \ldots, \phi_{ld_l}(t)]$, i.e.,*

$$\mathcal{S}_l \triangleq \left\{ X(t) | X(t) = \sum_{q=1}^{d_l} \alpha_q \phi_{lq}(t), \alpha_q \in \mathcal{R} \right\}. \tag{2}$$

*In this manuscript, we consider orthogonal basis functions, i.e., $\int_{\mathcal{T}} \phi_{lq}(t)\phi_{lm}(t)\mathrm{d}t = 0, \forall q, m = 1, \ldots, d_l, q \neq m$.*

In particular, we assume the subspace that $X_{ij}(t)$ belongs to is consistent for all the samples $i = 1, \cdots, N$, but their basis coefficients $\boldsymbol{\alpha}_{ij} = [\alpha_{ij1}, \ldots, \alpha_{ijd_l}]$ are different, i.e., sampled at random from the unit sphere in $\mathcal{R}^{d_l \times 1}$ for different samples. In addition, denote that $\mathcal{X}_l = \{X_j(t) | X_j(t) \in \mathcal{S}_l, j = 1, \ldots, p\}$ with the cardinality $p_l$. We have $\sum_{l=1}^{L} p_l = p$.

**Assumption 2.** *((A2) Self-Expressive Assumption) If there are sufficient functions from each subspace, i.e., $p_l > d_l$ for $l = 1, \ldots, L$, and, for these $p_l$ functions, no $d_l$ functions are spanned on the same $d_l - 1$ basis functions, then $X_{ij}(t)$ is **self-expressive**, which means for every $X_{ij}(t) \in \mathcal{X}_l$, we have*

$$X_{ij}(t) = \sum_{X_{ir}(t) \in \mathcal{X}_l, r \neq j} b_{jr} X_{ir}(t). \tag{3}$$

*This means that if $X_{ij}(t)$ is a function in $\mathcal{X}_l$, it can be represented as a linear combination of the other $p_l - 1$ functions in the same subspace. (It should be noted that the above equation is a general formula, and it does not mean that all $b_{jr}$ from the same subspace as $X_{ij}(t)$ should always have nonzero values.)*

With this assumption, $X_{ij}(t)$ can be recovered as a sparse solution of the multilinear regression equation $X_{ij}(t) = \mathbf{X}_i(t)\mathbf{b}_j$, with the regression coefficients $\mathbf{b}_j \in \mathcal{R}^{p \times 1}$ who has $b_{jr} \neq 0$ for $\{r | X_{ir}(t) \in \mathcal{X}_l, r \neq j\}$, and $b_{jr} = 0$ otherwise. Notably, for a system with equations such as (3), $\mathbf{b}_j$ may have infinite number of solutions, but we can restrict the set of solutions by minimizing an objective function with the $l_q$-norm of the solution, i.e.,

$$\min ||\mathbf{b}_j||_q, \tag{4}$$

subject to $X_{ij}(t) = \mathbf{X}_i(t)\mathbf{b}_j, b_{jj} = 0$.

Different choices of $q$ have different effects on the obtained solution. Typically, by decreasing the value of $q$ from infinity towards zero, the sparsity of the solution increases. The extreme case of $q = 0$ corresponds to the general NP-hard problem of finding the sparsest representation of the given function. Since we are interested in efficiently finding the non-trivial sparse representation of $X_{ij}(t)$ in the dictionary $\mathbf{X}_i(t)$, we consider minimizing the tightest convex relaxation of the $l_0$-norm, i.e., the $l_1$-norm, which can be solved efficiently using convex programming tools. Then we have the following theorem.

**Theorem 1.** *Under Assumptions (A1) and (A2), we define the following optimization problem*

$$\min_{\mathbf{b}_j} \lambda ||\mathbf{b}_j||_1 + \frac{1}{2} \sum_{i=1}^{N} \int_{\mathcal{T}} Z_{ij}(t) \Gamma_j(t,s)^{-1} Z_{ij}(s) \mathrm{d}t \mathrm{d}s, \tag{5}$$

$$subject\ to\ Z_{ij}(t) = Y_{ij}(t) - \mathbf{Y}_i(t)\mathbf{b}_j, \quad b_{jj} = 0,$$

*for $j = 1, \ldots, p$. It can be proven that under some regular conditions, if $\lambda$ is well tuned, with a very high probability, the solution of (5) will only have nonzero values for functions from the same subspace as $Y_{ij}(t)$.*

The detailed proof and the regular conditions of Theorem 1 can be referred in Wang and Xu (2013). Theorem 1 indicates that under regular conditions and proper choice of $\lambda$, with a high probability, solving (5) can recover the true subspace. With the obtained $\mathbf{b}_j, j = 1, \ldots, p$, define $\mathbf{B} = [\mathbf{b}_1, \ldots, \mathbf{b}_p] \in \mathcal{R}^{p \times p}$. It can be regarded as a representation of the cross-correlation structure of the $p$-dimensional functions. Actually, when $Y_{ij}(t)(t \in [0, T])$ degenerates to a scaler, i.e., $\mathbf{Y}_i$ becomes a $p$-dimensional vector, the estimated $b_{j_1 j_2}$ is exactly the sparse partial correlation between $Y_{ij_1}$ and $Y_{ij_2}$, and (5) degenerates to the graphical LASSO problem, which is usually used for sparse network modeling. Hereafter we denote the learning system with (5) as *static functional subspace learning(SFSL)*.

## 3.2 Dynamic Correlation Modeling via Fused LASSO

In this subsection, we consider the cross-correlations between different functions can change over time $t$, which is very common in reality. For many systems (as the example in Section 1), their cross-correlation structure generally remains constant for a certain time period, and changes to another constant state when the system undergoes some typically external disturbance. In another word, the cross-correlations only have stepwise changes at certain time points. Assume there are totally $S - 1$ change points $\tau_s(s = 1, \ldots, S - 1)$ inside $\mathcal{T}$ with totally $S$ time segments $\mathcal{T}^s(s = 1, \ldots, S)$. For every time segment, we have

$$Y_{ij}^s(t) = X_{ij}^s(t) + \epsilon_{ij}(t), t \in \mathcal{T}^s, \tag{6}$$

where $X_{ij}^s(t)$ is a function in the subspace $\mathcal{S}_l^s(l = 1, \ldots, L^s)$. For different segments $s = 1, \ldots, S$, the subspace that the $j^{th}$ function belongs to can change (but is consistent for all the samples), and the subspace number $L^s$ and structures $\mathcal{S}_l^s(l = 1, \ldots, L^s)$ can also be different. Similarly, denote $\mathcal{X}_l^s = \{X_{ij}^s | X_{ij}^s \in \mathcal{S}_l^s, j = 1, \ldots, p\}$ with the cardinality $p_l^s$. The distribution of $\epsilon_{ij}(t)$ is consistent for all the segments, without changes. In reality, the change points and changed subspace structures are usually unknown. To describe this dynamic system of (6) using a similar way as SFSL, $b_{jr}(j, r = 1, \ldots, p)$ should be dynamic over time, as $\mathbf{b}_j(t) = [b_{j1}(t), \ldots, b_{jp}(t)]'$ for $t \in \mathcal{T}$. Then $\mathbf{B}(t) = [\mathbf{b}_1(t), \ldots, \mathbf{b}_p(t)]', t \in \mathcal{T}$ would eventually capture the dynamic cross-correlations. However, this naive relaxation gives too much flexibility for the change of $\mathbf{b}_j$, which could lead to severe overfitting. With this in mind, to better regularize the dynamics, we consider penalizing the change of $\mathbf{b}_j(t)$, i.e., $d\mathbf{b}_j(t)/dt$, to encourage its local constancy by borrowing the idea from the fused LASSO (Tibshirani et al. 2005). In particular, we get $\mathbf{b}_j(t)$ as the solution of the following problem, i.e.

$$\min_{\mathbf{b}_j(t)} \lambda_1 \int_{\mathcal{T}} ||d\mathbf{b}_j(t)/dt||_1 dt + \lambda_2 \int_{\mathcal{T}} ||\mathbf{b}_j(t)||_1 dt + \frac{1}{2} \sum_{i=1}^{N} \int_{\mathcal{T}} Z_{ij}(t) \Gamma_j(t, s)^{-1} Z_{ij}(s) dt ds, \quad (7)$$

$$\text{subject to } Z_{ij}(t) = Y_{ij}(t) - \mathbf{Y}_i(t)\mathbf{b}_j(t), \quad b_{jj}(t) = 0, \quad t \in \mathcal{T},$$

for $j = 1, \ldots, p$, separately. In (7), the second term encourages sparsity in the regression coefficients with the tuning parameter $\lambda_2$. The first term encourages sparsity in their differences, i.e., the flatness of the coefficient functions $\mathbf{b}_j(t)$, with the tuning parameter $\lambda_1$.

As introduced earlier, in practice, $\mathbf{Y}_i(t)$ are usually recorded at the grid of discrete points. In this paper we assume that the grid points are dense and equally spaced at $n$ points, i.e., $\{t_k, 1 \leq k \leq n\}$ with $t_1 < \ldots < t_n$, then we have $\mathbf{Y}_i, \mathbf{X}_i \in \mathcal{R}^{n \times p}$ with every row $\mathbf{Y}_i(t_k), \mathbf{X}_i(t_k) \in \mathcal{R}^{1 \times p}$ and every column $\mathbf{Y}_{ij}, \mathbf{X}_{ij}$. In this manuscript, we tackle the normalized model with $||\mathbf{X}_{ij}||_2 = 1$. With the signal-to-noise assumption, we have $\boldsymbol{\epsilon}_{ij} \in \mathcal{R}^{n \times 1}$ with $\mathrm{E}[\epsilon_{ij}(t)] = 0$, $\mathrm{Var}\,[\epsilon_{ij}(t)] = \sigma^2/n$ and $\mathrm{E}[\boldsymbol{\epsilon}_{ij}\boldsymbol{\epsilon}_{ij}']/\mathrm{Var}[\epsilon_{ij}(t)] = \mathbf{\Gamma}_j$. Then (7) can be reformulated in terms of $b_{jr}(t_k)$ for $r = 1, \ldots, p, r \neq j, k = 1, \ldots, n$ as

$$\min_{b_{jr}(t_k), r \neq j, k=1,\ldots,n} \lambda_1 \sum_{k=2}^{n} ||\mathbf{b}_j(t_k) - \mathbf{b}_j(t_{k-1})||_1 + \lambda_2 \sum_{k=1}^{n} ||\mathbf{b}_j(t_k)||_1 + \frac{1}{2} \sum_{i=1}^{N} \mathbf{Z}_{ij}' \mathbf{\Gamma}_j^{-1} \mathbf{Z}_{ij}, \quad (8)$$

$$\text{subject to } Z_{ij}(t_k) = Y_{ij}(t_k) - \mathbf{Y}_i(t_k)\mathbf{b}_j(t_k), \quad b_{jj}(t_k) = 0, \text{ for } k = 1, \ldots, n.$$

Hereafter, we name the learning system with (8) as *dynamic functional subspace learning (DFSL)*. Based on Equation (8), we can estimate the change points for function $Y_{ij}(t)$ as those $\{\hat{\tau}_s, s = 1, \ldots, \hat{S} - 1 | b_{jr}(\hat{\tau}_s) - b_{jr}(\hat{\tau}_s - 1) \neq 0, \exists r = 1, \ldots, p\}$ where $\hat{S} - 1$ is the number of estimated change points ($\hat{S}$ is the number of estimated segments).

## 3.3 Theoretical Properties

Now we detail the assumptions of the proposed DFSL, under which its theoretical properties can be better established. In particular, it is assumed that there are $S - 1$ cross-correlation change points $1 < \tau_1 < \ldots < \tau_{S-1} < n$ in the $n$-length functional data with $\tau_0 = 1$ and $\tau_S = n + 1$. For every time segment $\mathcal{T}^s = [\tau_{s-1}, \tau_s)$, the true constant cross-correlations can be represented by $[\boldsymbol{\beta}_{1,s}, \ldots, \boldsymbol{\beta}_{p,s}]$ where $\boldsymbol{\beta}_{j,s} = \mathbf{b}_j(t_k), k \in \mathcal{T}^s$. Except for Assumptions (A1)-(A2), we listed the additional assumptions here.

**Other Assumptions.**

*(A3) The sequence $\{\delta_N\}_{n \geq 1}$ is a non-increasing and positive sequence tending to zero as $N$ tends to infinity and satisfies $N \delta_N (\xi_{min})^2 / \log(N) \to \infty$.*

*(A4) The minimum interval length $\Delta_{min} = \min_{1 \leq s \leq S-1} |\tau_{s+1} - \tau_s|$ satisfies $\Delta_{min} \geq \delta_N$.*

*(A5) The minimum coefficient change $\xi_{min} = \min_{\substack{1 \leq s \leq S-1 \\ j=1,\ldots,p}} ||\boldsymbol{\beta}_{j,s+1} - \boldsymbol{\beta}_{j,s}||_2$ has a lower bound.*

*(A6) The maximum coefficient change $\xi_{max} = \max_{\substack{1 \leq s \leq S-1 \\ j=1,\ldots,p}} ||\boldsymbol{\beta}_{j,s+1} - \boldsymbol{\beta}_{j,s}||_2$ has a upper bound.*

*(A7) The noise $\boldsymbol{\epsilon}_{ij}$ follows the normal distribution with mean $\mathbf{0}$ and covariance matrix $\sigma^2 \boldsymbol{\Gamma}_j / n$.*

*(A8) The affinity between two subspaces $\mathcal{S}_l$ and $\mathcal{S}_r$ is defined as $||\boldsymbol{\Phi}_l' \boldsymbol{\Phi}_r||_F$ where $\boldsymbol{\Phi}_l$ and $\boldsymbol{\Phi}_r$ are the orthogonal basis functions of $\mathcal{S}_l$ and $\mathcal{S}_r$ respectively, and $|| \cdot ||_F$ is the Frobenius norm of the matrix. We assume*

$$\max_{\substack{s=1,\ldots,S \\ l,r=1,\ldots,L^s}} \frac{\mathit{aff}(\mathcal{S}_l^s, \mathcal{S}_r^s)}{\sqrt{\max(d_l^s, d_r^s)}} \leq \kappa_0 / \log p,$$

*where $\kappa_0$ is a constant. This assumption states that any two subspaces cannot be too close to each other, which is necessary and sufficient for correct identification of functions of every subspace.*

**Theorem 2.** *Under Assumptions (A1)-(A7), the change-points $\{\hat{\tau}_s, s = 1, \ldots, \hat{S} - 1\}$ estimated by (8) where $\hat{S}$ is the number of estimated time segments, satisfy that, if $\hat{S} = S$, with probability tending to one:*

$$P(\max_{1 \le s \le S-1} |\hat{\tau}_s - \tau_s| \le \delta_N) \to 1, \ as \ N \to \infty, \tag{9}$$

*when $\lambda_1$ and $\lambda_2$ satisfy $(N\delta_N \xi_{min})^{-1}\lambda_1 \to 0$ and $(N\xi_{min})^{-1}\lambda_2 \to 0$.*

The proof of Theorem 2 is given in the Appendix. Theorem 2 states that when the number of segments is known or correctly estimated, the estimated segments recover the true ones consistently as the time length increases.

**Theorem 3.** *Based on Theorem 2, assume Assumption (A8) also holds. Suppose $Y_{ij}^s \in \mathcal{X}_{l_s}^s$ for time segment $\mathcal{T}^s$. We have as $N \to \infty$, the probability that the solution of (8) will only have nonzero values for functions from the same subspace as $Y_{ij}^s(t_k)$ in every time segment $\mathcal{T}^s$ is at least $1 - c_1/p_l^2 - c_2 \exp(-\Delta_{min}\omega^2/8)$. Here $\omega$ is a positive constant satisfies*

$$\sqrt{1+\omega} \le \frac{1 + \sigma^2 - \sqrt{24}\log p(\max_{\substack{r,l=1,\ldots,L^s \\ s=1,\ldots,S}} \frac{aff(\mathcal{S}_r^s, \mathcal{S}_l^s)}{\sqrt{\max(d_r^s, d_l^s)}}) - 2\sigma}{\sqrt{24}\frac{\log p}{\sqrt{p}}\sigma + 2\sigma^2},$$

*and $\lambda_1/\lambda_2 \to 0$ and $\lambda_2/\sqrt{N\log N} \to \infty$ are also satisfied as $N \to \infty$. $c_1$ and $c_2$ are two fixed numerical constants.*

The proof of Theorem 3 is provided in the Appendix. Theorem 3 states that when $\Delta_{min}$ and $p_l$ increase, as long as the subspaces are not too close with each other, DFSL can separate functions from different subspaces correctly. This point will be further demonstrated in Section 5.

# 4  Model Inference

In this section, we will first introduce an efficient estimation method for the proposed DFSL model in 4.1. Then we will talk about its tuning parameter selection in 4.2. Based on the estimated DFSL model, we will discuss subspace inference in 4.3.

## 4.1 Optimization via Fast Iterative Shrinkage-thresholding Algorithm

(8) is a convex problem including two parts: the smooth part for $\mathbf{b}_j$, i.e.,

$$f(\mathbf{b}_j) = \frac{1}{2} \sum_{i=1}^{N} \mathbf{Z}'_{ij} \mathbf{\Gamma}_j^{-1} \mathbf{Z}_{ij},$$

and the non-smooth part for $\mathbf{b}_j$, i.e.,

$$g(\mathbf{b}_j) = \lambda_1 \sum_{k=1}^{n} ||\mathbf{b}_j(t_k)||_1 + \lambda_2 \sum_{k=2}^{n} ||\mathbf{b}_j(t_k) - \mathbf{b}_j(t_{k-1})||_1.$$

Several categories of first-order methods have been developed to optimize this kind of composite function. Among them, the most popular one is in the class of iterative shrinkage-thresholding algorithm (ISTA). As one of the proximal gradient methods, the basic idea of ISTA is to build an approximation model for the composite function with a regularization at each iteration. ISTA has a convergence rate of $O(1/k)$, where $k$ is the number of iterations. To achieve a faster convergence rate, Beck and Teboulle (2009) extended the Nesterov's method (Nemirovski 2005) and proposed the fast ISTA (FISTA) algorithm. The main difference between ISTA and FISTA is that the later employs the approximation model not only based on the previous iteration, but rather on the previous two iterations. FISTA generally has the convergence rate of $O(1/k^2)$. Liu et al. (2010) used it to solve the one-dimensional fused LASSO problem. Xin et al. (2016) also applied it in the generalized fussed LASSO estimation. In light of its fast convergence rate and simplicity, here we propose to use FISTA in Beck and Teboulle (2009) to solve (8).

The pillar of ISTA-based methods is to construct the following model to approximate the composite objective function based on a searching point $\mathbf{s}_j$ as

$$Q_{L_j}(\mathbf{b}_j, \mathbf{s}_j) := f(\mathbf{s}_j) + \langle \mathbf{b}_j - \mathbf{s}_j, \nabla f(\mathbf{s}_j) \rangle + \frac{L_j}{2} ||\mathbf{b}_j - \mathbf{s}_j||^2 + g(\mathbf{b}_j), \qquad (10)$$

where $L_j$ is a constant bigger than the Lipschitz constant of $\nabla f$. With (10), we can develop the following gradient descent-like method for solving (8) by iteratively minimizing (10), i.e.,

$$\mathbf{b}_j^{k+1} = \arg\min_{\mathbf{b}_j} Q_{L_j}(\mathbf{b}_j, \mathbf{s}_j^k), \tag{11}$$

where $k$ is the iteration step, and $\mathbf{s}_j^k$ is the searching point for the current step.

**Proposition 1.** *For the objective function of (8), (10) and (11) can be derived and decomposed for every* $\mathbf{b}_{jr} = [b_{jr}(t_1), \ldots, b_{jr}(t_n)]$ *for* $r = 1, \ldots, p, r \neq j$ *separately, as*

$$\mathbf{b}_{jr}^{k+1} = \min_{\mathbf{b}_{jr}} \frac{1}{2}||\mathbf{b}_{jr} - \mathbf{z}_{jr}||_2^2 + s_{j1}||\mathbf{b}_{jr}||_1 + s_{j2}||\mathbf{D}\mathbf{b}_{jr}||_1, \tag{12}$$

*where* $\mathbf{z}_{jr} = \mathbf{s}_{jr}^k - \frac{1}{L_j}\sum_{i=1}^N diag(\tilde{\mathbf{Y}}_{ij})(\tilde{\mathbf{Y}}_{ij} - \sum_{r \neq j} diag(\tilde{\mathbf{Y}}_{ir})\mathbf{s}_{jr}^k)$ *with* $\tilde{\mathbf{Y}}_{ij} = \boldsymbol{\Gamma}_j^{-\frac{1}{2}}\mathbf{Y}_{ij}$ *for* $j = 1, \ldots, p, i = 1, \ldots, N$; $\mathbf{D} \in \mathcal{R}^{(n-1) \times n}$ *is the first-order difference matrix whose* $D_{kk} = -1, D_{k(k+1)} = 1$ *for* $k = 1, \ldots, n-1$, *and other components equal* $0$.

(12) is the exact form of the fused LASSO signal appropriator (FLSA) function with $L_j = ||\boldsymbol{\Upsilon}_{-j}||_2^2$ where $\boldsymbol{\Upsilon}_{-j} = \left[\boldsymbol{\Upsilon}'_{1(-j)}, \ldots, \boldsymbol{\Upsilon}'_{N(-j)}\right]'$ and $\boldsymbol{\Upsilon}_{i(-j)} = [diag(\tilde{\mathbf{Y}}_{i1}), \ldots, diag(\tilde{\mathbf{Y}}_{i(j-1)}), diag(\tilde{\mathbf{Y}}_{i(j+1)}), \ldots, diag(\tilde{\mathbf{Y}}_{ip})]$, for $i = 1, \ldots, N$. $s_{j1}$ and $s_{j2}$ are scaled $\lambda_1$ and $\lambda_2$ by $L_j$. Consequently, Proposition 1 indicates that the objective function of (8) could be decomposed and solved efficiently using the FLSA estimators. Here we adopt the method in Liu et al. (2010).

In FISTA, the searching point $\mathbf{s}_{jr}^k$ in every iteration is defined as

$$\mathbf{s}_{jr}^k = \mathbf{b}_{jr}^k + \rho_k(\mathbf{b}_{jr}^k - \mathbf{b}_{jr}^{k-1}),$$

where $\rho_k = (\sqrt{1 + 4\rho_{k-1}^2})/2$ is an iteratively chosen coefficient. The detailed algorithm of solving (8) based on FISTA is shown in Algorithm 1.

**Remark 1.** *In practice,* $\{\Gamma_j(t, s), \sigma\}$ *may be unknown.* $\sigma$ *may be even relaxed to be different for* $j = 1, \ldots, p$. *In this case we need to estimate* $\{\boldsymbol{\Gamma}_j, \sigma_j\}$ *as well. Then (8) becomes no longer convex. As such, we may use the block coordinate descent (BCD) algorithm to estimate* $\mathbf{b}_{jr}(r = 1, \ldots, p, r \neq j)$ *and* $\boldsymbol{\Sigma}_j = \sigma_j^2 \boldsymbol{\Gamma}_j/n$ *separately and iteratively, for* $j = 1, \ldots, p$. *The detailed algorithm is shown in Algorithm 2.*

**Data:** $\mathbf{Y}_i, i = 1, \ldots, N, \rho_0, \mathbf{\Gamma}_j, L_j, \mathbf{b}_{jr}^0, r = 1, \ldots, p, r \neq j$

**Result:** Estimated $\mathbf{b}_{jr} = \mathbf{b}_{jr}^{k+1}, r = 1, \ldots, p, r \neq j$

**initialization**

Initialize $\mathbf{b}_{jr}^1 = \mathbf{b}_{jr}^0$ for $r = 1, \ldots, p, r \neq j$

**Estimation**

**while** $\sum_{r \neq j} ||\mathbf{b}_{jr}^k - \mathbf{b}_{jr}^{k-1}||_2^2 > e_1$ **do**

    Set $\rho_k = (\sqrt{1 + 4\rho_{k-1}^2})/2$

    Set $\mathbf{s}_{jr}^k = \mathbf{b}_{jr}^{k-1} + \rho_k(\mathbf{b}_{jr}^k - \mathbf{b}_{jr}^{k-1})$ for $r = 1, \ldots, p, r \neq j$

    **for** $r = 1, \ldots, j-1, j+1, \ldots, p$ **do**

        $\mathbf{z}_{jr} = \mathbf{s}_{jr}^k - \frac{1}{L_j} \sum_{i=1}^N \mathrm{diag}(\tilde{\mathbf{Y}}_{ij}) \left[ \tilde{\mathbf{Y}}_{ij} - \sum_{r \neq j} \mathrm{diag}(\tilde{\mathbf{Y}}_{ir})\mathbf{s}_{jr}^k \right]$

        Solve $\mathbf{b}_{jr}^{k+1} = \min_{\mathbf{b}_{jr}} \frac{1}{2}||\mathbf{b}_{jr} - \mathbf{z}_{jr}||_2^2 + s_{j1}||\mathbf{b}_{jr}||_1 + s_{j2}||\mathbf{D}\mathbf{b}_{jr}||_1$ using FLSA

    **end**

    Set $k = k + 1$

**end**

**Algorithm 1:** Solving $\mathbf{b}_{jr}(r = 1, ..., p, r \neq j)$ based on FISTA

## 4.2 Tuning Parameter Selection

In general, the selection of optimal tuning parameters for a given model can be a difficult task, which is further complicated as the number of tuning parameters increases. Here we propose to follow the tuning procedure for the fused LASSO in Nowak et al. (2011) to reduce the computation. Specifically, to simplify the search for the optimal tuning parameters, given the sample size $N$, we reparameterize $\lambda_1$ and $\lambda_2$ in terms of $\lambda_0$ and $\rho \in (0, 1)$ (here without confusion, we omit $N$ in the later notation for conciseness), such that $\lambda_1 = \rho \lambda_0$ and $\lambda_2 = (1 - \rho)\lambda_0$. We can think of $\lambda_0$ as an overall tuning parameter with $\rho$ determining how much emphasis is placed on sparsity versus smoothness. By fixing the possible values that $\rho$ can take, we effectively reduce the search over $\lambda_1$ and $\lambda_2$, to a search over one parameter $\lambda_0$. In particular, we initially fix the possible values of $\rho$ (e.g. $\{0.1, 0.3, 0.5, 0.7, 0.9\}$). For each value of $\rho$, we find the value of $\lambda_0$ that results in each estimated variable to be 0, and denote this value by $\lambda_{0,\rho}^{\max}$. Then we chose a fixed number of candidate values for $\lambda_0$ from the interval $(0, \lambda_{0,\rho}^{\max})$. The optimal values of $\rho$ and $\lambda_0$ are selected by searching over this 2D grid for the value that minimizes the following criterion:

$$(Nn) \sum_{j=1}^p \log \left( \frac{\sum_{i=1}^N \mathbf{Z}_{ij}' \mathbf{\Gamma}_j^{-1} \mathbf{Z}_{ij}}{Nn} \right) + \log(Nn) \sum_{j=1}^p k_{\rho,\lambda_0}(j). \tag{13}$$

**Data:** $\mathbf{Y}_i, i = 1, \ldots, N, \rho_0$

**Result:** Estimated $\mathbf{b}_j^g, \mathbf{\Sigma}_j^g, j = 1, \ldots, p$

***Initialization*** *(by the least square method for the ordinary linear regression)*

Set $\mathbf{\Gamma}_j^0 = \mathbf{I}_n$ for $j = 1, \ldots, p$

**for** $j = 1, \ldots, p$ **do**

    Initialize $\mathbf{b}_j^0(t_k)$ by the multivariate linear regression for $k = 1, \ldots, n$ as

    **for** $k = 1, \ldots, n$ **do**

        $\mathbf{b}_j^0(t_k)' = (\sum_{i=1}^N \tilde{\mathbf{Y}}_{i(-j)}(t_k)' \tilde{\mathbf{Y}}_{i(-j)}(t_k))^{-1}(\sum_{i=1}^N \tilde{\mathbf{Y}}_{i(-j)}(t_k)' \tilde{Y}_{ij}(t_k))$

    **end**

**end**

***Estimation*** *(by the block coordinate descent method)*

**while** $\sum_{j=1}^p ||\mathbf{b}_j^g - \mathbf{b}_j^{g-1}||_2^2 > e_2$ *or* $\sum_{j=1}^p ||\mathbf{\Sigma}_j^g - \mathbf{\Sigma}_j^{g-1}||_2^2 > e_3$ **do**

    **for** $j = 1, \ldots, p$ **do**

        Calculate the residual of $\mathbf{Y}_{ij}, j = 1, \ldots, p$ as

        **for** $k = 1, \ldots, n$ **do**

            Estimate $\epsilon_{ij}(t_k)$ for every $i = 1, \ldots, N$ as

$$\epsilon_{ij}(t_k) = Y_{ij}(t_k) - \sum_{r \neq j} Y_{ir}(t_k) b_{jr}^g(t_k)$$

        **end**

        Set $\mathbf{\Sigma}_j^g = \sum_{i=1}^N \boldsymbol{\epsilon}_{ij}' \boldsymbol{\epsilon}_{ij}$

        Estimate $\mathbf{b}_j^g$ using FISTA, with input values as

        $\mathbf{\Gamma}_j = \mathbf{\Sigma}_j^g, \mathbf{b}_{jr}^0 = \mathbf{b}_{jr}^{g-1}, r = 1, \ldots, p, r \neq j$

    **end**

    Set $g = g + 1$

**end**

**Algorithm 2:** Estimate $\mathbf{b}_{jr}(r = 1, \ldots, p, r \neq j)$ and $\mathbf{\Sigma}_j$ for $j = 1, \ldots, p$ based on BCD

Here $k_{\rho,\lambda_0}(j)$ is the number of nonzero elements in $\mathbf{b}_j$ given the current $\rho$ and $\lambda_0$. The term of $\sum_{j=1}^{p} k_{\rho,\lambda_0}(j)$ represents the complexity of the model, with larger values indicating greater complexity. This criterion is similar to the Bayesian information criterion. Its rational is that by minimizing (13), we attempt to find an appropriate model without overfitting the data. The first term will tend to be smaller for complex models, whereas the second term will tend to be smaller for simple models. For computational reasons, we prefer this approach for selecting the optimal tuning parameters.

## 4.3 Identifying the Change Point and Latent Basis Functions

Based on the estimated model, we first discuss how to identify the cross-correlation change points in Section 4.3.1, and then discuss how to estimate the subspace structure and its corresponding basis functions for every time segment in Section 4.3.2.

### 4.3.1 Identifying the Change Point

In particular, we define

$$c_{jk} = \sum_{r \neq j} |\hat{b}_{jr}(t_k) - \hat{b}_{jr}(t_{k-1})|, k = 2, \ldots, n, j = 1, \ldots, p, \tag{14}$$

where $\hat{b}_{jr}(t_k)$ is the solution of (8) solved by Algorithm 1 or Algorithm 2. (14) is used to identify the potential change points for function $\mathbf{X}_j(t)$ by testing whether $c_{jk}$ is bigger than a threshold $c_{j0}$, i.e., $T_j = \{k | c_{jk} > c_{j0}\}$. The threshold depends on the pre-specified system sensitivity. Then for every time point $t_k$, we calculate $C_k = \sum_{j=1}^{p} \mathcal{I}(t_k \in T_j)$, and use $C_k (1 \leq k \leq n)$ for system-level change point decision-making.

### 4.3.2 Identifying the Latent Subspaces and Basis Functions

Suppose we have identified totally $S-1$ change points $\hat{\tau}_s (s = 1, \ldots, S-1)$. For every segment, we calculate the average DFSL coefficient $\bar{\mathbf{b}}_j^s \in \mathcal{R}^{p \times 1}$ where $\bar{b}_{jr}^s = \sum_{l=\tau_{s-1}+1}^{\tau_s} |\hat{b}_{jr}(t_k)|$ for $r \neq j$ and $\bar{b}_{jj}^s = 0$. Define $\bar{\mathbf{B}}^s = [\bar{\mathbf{b}}_1^s, \ldots, \bar{\mathbf{b}}_p^s]$. Then similar to the procedure of Elhamifar and Vidal (2009); Bahadori et al. (2015), we can define the symmetric affinity matrix $\mathbf{A}^s = \bar{\mathbf{B}}^s + \bar{\mathbf{B}}^{s\prime}$, and apply the spectral clustering (Ng et al. 2001) on $\mathbf{A}^s$ for subspace clustering (When the number of subspaces is unknown priorly, other clustering methods, such as hierarchical clustering, can be applied as well). Notably, in our model we allow the number of subspaces or the subspace basis functions to change for different time segments.

Suppose for the segment $s$, there are $L^s$ subspaces identified, and the $l^{\text{th}}$ subspace $\mathcal{S}_l^s$ has the cardinality $p_l^s$ and the function set $\mathbf{Y}_{i(l)}^s = \{\mathbf{Y}_{i1}^s, \ldots, \mathbf{Y}_{ip_l^s}^s\}$, where $\mathbf{Y}_{ij}^s$ are the signals of $\mathbf{Y}_{ij}$ in the $s^{\text{th}}$ time segment, i.e., $Y_{ij}^s(t_k)(k = \tau_{s-1}, \ldots, \tau_s - 1, j = 1, \ldots, p_l^s)$. Then we can estimate the basis functions of $\mathcal{S}_l^s$ using the multi-channel functional PCA (MFPCA) (Paynabar et al. 2016). However, since here we assume that the basis functions are smooth, we optimize the loss function of MFPCA together with a smoothness regularization. In particular, our objective function is

$$\min_{\boldsymbol{\alpha}_{iq}^s, \phi_{lq}^s, q=1,\ldots,d_l^s} \sum_{i=1}^{N} ||\mathbf{Y}_{i(l)}^s - \sum_{q=1}^{d_l^s} \phi_{lq}^s \boldsymbol{\alpha}_{iq}^{s\prime}||_F^2 + \lambda_3 \sum_{q=1}^{d_l^s} \sum_{k=2}^{n_s} (\phi_{lq}^s(t_k) - \phi_{lq}^s(t_{k-1}))^2, \tag{15}$$

$$\text{subject to } ||\phi_{lq}^s||_2 = 1, \phi_{lq}^{s\prime} \phi_{lr}^s = 0, \forall q, r = 1, \ldots, d_l^s, q \neq r,$$

where $\phi_{lq}^s \in \mathcal{R}^{n_s \times 1}$ with $n_s = \tau_s - \tau_{s-1}$ is the $q^{\text{th}}$ basis function, and $\boldsymbol{\alpha}_{iq}^s \in \mathcal{R}^{p_l^s \times 1}$ are the projections of $\mathbf{Y}_{i(l)}^s$ on $\phi_{lq}^s$. We chose $d_l^s$ such that the explained cumulative percentage of the sample variance by the first $d_l^s$ MFPCA loadings is 95%. Following a similar way as Huang et al. (2008), (15) can be efficiently solved.

# 5 Numerical Studies

In this section, to evaluate the effectiveness of DFSL, we perform some numerical experiments using synthetic data generated from the assumed subspace model described in Section 3.2. We will first illustrate the description power of DFSL for multivariate functional data and the efficiency of the proposed estimation algorithm. Then we will compare DFSL with some state-of-art methods. Finally, we will talk about the sensitivity analysis of DFSL.

## 5.1 Synthetic Data Experiments

We assume every function sample has total $n$ equally spaced sampling time points. Among them there are $S - 1$ correlation change points $\tau_s(s = 1, \ldots, S - 1)$ with a total of $S$ time segments, i.e., $n = \sum_{s=1}^{S} n_s$. For every time segment, we assume that there exist $L^s$ subspaces $\mathcal{S}_l^s (l = 1, \ldots, L^s)$. Then for every sample $i$, we can denote its functional data in the $s^{\text{th}}$ time segment as $\mathbf{Y}_i^s = \left[ \mathbf{Y}_{i(1)}^s, \ldots, \mathbf{Y}_{i(L^s)}^s \right]$ with the $l^{\text{th}}$ subspace data $\mathbf{Y}_{i(l)}^s \in \mathcal{R}^{n_s \times p_l^s}$. We propose to generate

$$\mathbf{Y}_{i(l)}^s = \mathbf{\Phi}_l^s \mathbf{A}_{i(l)}^s + \mathbf{E}_{i(l)}^s, \quad l = 1, \ldots, L^s,$$

for $s = 1, \ldots, S$. $\mathbf{E}_{i(l)}^s = \left[ \boldsymbol{\epsilon}_{i1}^s, \ldots, \boldsymbol{\epsilon}_{ip_l^s}^s \right]$ is the noise function. Here we assume that the noise is also time-segmented with the autocorrelation matrix $\mathbf{\Gamma}_j^s$. $\mathbf{\Phi}_i^s \in \mathcal{R}^{n_s \times d_l^s}$ are the basis functions for $\mathcal{S}_l^s$, and $\mathbf{A}_{i(l)}^s \in \mathcal{R}^{d_l^s \times p_l^s}$ is the weight matrix with the $(q, j)$ component as the weight of the $q^{\text{th}}$ basis function $\phi_{lq}^s$ on the $j^{\text{th}}$ function. We further consider that $\mathbf{A}_{i(l)}^s$ can be decomposed into two parts, i.e.,

$$\mathbf{A}_{i(l)}^s = \mathbf{R}_{i(l)}^s \mathbf{V}_l^s,$$

where $\mathbf{V}_l^s \in \mathcal{R}^{m_l^s \times p_l^s}$ is the variation matrix, with every row orthogonal to each other indicating one variation pattern; $\mathbf{R}_{i(l)}^s \in \mathcal{R}^{d_l^s \times m_l^s}$ is the variation coefficient matrix, with every row $\mathbf{r}_{iq}^s (q = 1, \ldots, d_l^s)$. Here in our simulation below, $p_l^s$, $d_l^s$, $m_l^s$, $L^s$ and the subspace that the $\mathbf{Y}_{ij} (j = 1, \ldots, p)$ belongs to are temporarily assumed to be unchanged for different time segments. Hence we omit the superscript $s$ hereafter (However, in practice, these parameters may vary as the case studies show in Section 6). Then we consider the following two models.

- Model (I): We have $p = 8$ functions. The total time length of every function is $n = 40$ with a cross-correlation change point $\tau_1 = 21$. Then we have $S = 2$ time segments with equal length, i.e., $n_1 = n_2 = 20$. For every segment, we assume that the $p$ functions come from $L = 2$ subspaces. The first four functions come from the B-spline space with $d_1 = 3$, i.e., $\phi_{1q}^s (q = 1, 2, 3)$ are the $1^{\text{st}}$, $4^{\text{th}}$, and $7^{\text{th}}$ B-spline basis functions of order 3 with a grid of $n_s$ equally spaced knots in $[0, 1]$ for $s = 1, 2$ respectively. The second four functions come from the Fourier space with $d_2 = 3$, i.e., $\phi_{2q}^s = \cos(qt_k + q\pi)(q = 1, 2, 3)$ with a grid of $n_s$ equally spaced time points $t_1, \ldots, t_{n_s}$ in $[0, 2\pi]$ for $s = 1, 2$ respectively. We assume $m_l = 2$ with $\mathbf{r}_{iq}^s$ following a two-dimensional multivariate normal distribution with mean vector $\mathbf{0}$ and the same covariance matrix $\mathbf{\Sigma}$ whose $(\mathbf{\Sigma})_{uv} = 0.5^{|u-v|}(u, v = 1, 2)$, for $q = 1, 2, 3, l = 1, 2$, and $s = 1, 2$. For $\mathbf{\Gamma}_j^s$, we set it the same for all $j = 1, \ldots, p$ with $(\mathbf{\Gamma}^s)_{uv} = 0.2^{|u-v|}(u, v = 1, \ldots, n_s)$ for $s = 1, 2$.

- Model (II): We have $p = 12$ functions. The total time length of every function is $n = 128$ with two cross-correlation change points, i.e., $\tau_1 = 33$ and $\tau_2 = 65$. Then we have $S = 3$ segments, i.e., $n_1 = n_2 = 32$ and $n_3 = 64$. For every segment, we assume that the $p$ functions come from $L = 3$ subspaces. The first four functions come from the B-spline space with $d_1 = 3$, i.e., $\phi_{1q}^s (q = 1, 2, 3)$ are the 1$^{\text{st}}$, 4$^{\text{th}}$ and 7$^{\text{th}}$ B-spline basis functions of order 3 with a grid of $n_s$ equally spaced knots in $[0, 1]$ for $s = 1, 2, 3$ respectively. The second four functions come from the Fourier space with $d_2 = 3$, i.e., $\phi_{2q}^s = \cos(qt_k + q\pi)(q = 1, 2, 3)$ with a grid of $n_s$ equally spaced time points $t_1, \ldots, t_{n_s}$ in $[0, 2\pi]$ for $s = 1, 2, 3$ respectively. The last four functions come from the wavelet space, with $d_3 = 3$, i.e., $\phi_{3q}^s(q = 1, 2, 3)$ are the $q^{\text{th}}$ Vaidyanathan basis functions. We assume $m_l = 2$ with $\mathbf{r}_{iq}^s$ generated in the same way as Model (I) for $q = 1, 2, 3, l = 1, 2, 3$ and $s = 1, 2, 3$. $\boldsymbol{\Gamma}_j^s$ is also the same as Model (I) for $j = 1, \ldots, p, s = 1, 2, 3$.

For each model, we generate 500 samples $\mathbf{Y}_i(i = 1, \ldots, 500)$ from the corresponding model with the noise standard deviation $\sigma = 0.05$, and use them to estimate $\mathbf{b}_j$ and $\boldsymbol{\Gamma}_j(j = 1, \ldots, p)$ for DFSL. As shown in Figure 6, for every estimated $\hat{\mathbf{b}}_j$, only the functions in the same subspace as $\mathbf{Y}_{ij}$ have nonzero regression coefficients (such as the first four functions from the B-spline subspace and the second four functions from the Fourier subspace). They keep constant over time and only have a step-wise change at around $t_{21}$. The estimated $\hat{\boldsymbol{\Sigma}}_j$ for Model (I) are jointly shown in Figure 7. Clearly, $\hat{\boldsymbol{\Sigma}}_j$ approximates to the true $\boldsymbol{\Sigma}_j = \text{diag}(\boldsymbol{\Sigma}_j^1, \boldsymbol{\Sigma}_j^2)$ with a standardized norm error $||\hat{\boldsymbol{\Sigma}}_j - \boldsymbol{\Sigma}_j||_2/||\boldsymbol{\Sigma}_j||_2 = 0.02$, demonstrating the efficiency of the proposed estimation method. Similar results also appear for Model (II). We further demonstrate the description power of DFSL by evaluating its estimation curve for one test sample. For comparison, we also estimate and plot the estimation results of SFSL. As shown in Figure 8, DFSL can predict the multivariate functions very well, while unsurprisingly SFSL fails since it cannot capture the dynamic cross-correlations.

Based on the estimated model, we further use the smooth MFPCA for subspace inference, and compare the extracted basis functions $\hat{\boldsymbol{\Phi}}_l^s$ with the true ones $\boldsymbol{\Phi}_l^s$. However, it should be noted that though the extracted PCA loadings can capture the subspace features, $\hat{\boldsymbol{\Phi}}_l^s$ may look a bit different from $\boldsymbol{\Phi}_l^s$, since the prior may be a rotation of the later. To eliminate the rotation influence, we rotate the extracted PCA loadings back using a rotation matrix $\mathbf{R}$ which minimizes the $||\mathbf{R}\hat{\boldsymbol{\Phi}}_l^s - \boldsymbol{\Phi}_l^s||^2$. Then the extracted basis functions after rotation, i.e., $\mathbf{R}\hat{\boldsymbol{\Phi}}_l^s$, for the three subspaces in the time segment $t \in [t_{65}, t_{128}]$ of Model (II) are shown in Figure 9. They match the true basis functions very well.

## 5.2 Comparison with State of the Arts

To better evaluate the performance of DFSL, besides SFSL, we have also made comparison of DFSL with some other state-of-the-art models introduced in Section 1, including the multi-channel functional PCA of Paynabar et al. (2016) (denoted as MFPCA), the functional PCA-based sparse graph model of Qiao et al. (2015) (denoted as FPCAGra), and the kernel based graphic LASSO of Kolar and Xing (2011) (denoted as KerGra).

We first compare their description power for multivariate functional data with dynamic cross-correlations. In particular, for every simulation replication, we generate 500 samples from Model (I) with a certain noise standard devision $\sigma$, and fit the data using these five models. In particular, for SFSL, we solve (5) using the LASSO algorithm with the penalty parameter tuned by the BIC. For MFPCA and FPCAGra, we estimate the corresponding PCA loadings and the distributions of the scores using the methods in their paper and determine the number of PCs by ensuring to capture 95% percentage of sample variation. For KerGra, it does not model the data but directly analyzes the function partial correlations using kernel smoothing methods. Hence here we tune its kernel parameter to ensure its smoothing leads to a comparable MSE as DFSL, and use this parameter for later partial correlation estimation and performance evaluation.

We test their modeling power by calculating their estimation mean square error (MSE) for additional 50 samples. The results based on 100 replications for different noise magnitude are shown in Figure 11a. As expected, the MSE of those five models increase with the noise magnitude. When the noise standard deviation grows close to the magnitude of the signal, i.e., $\sigma = 0.5$, all the models lose their inference accuracy. Among these models, DFSL consistently has the smallest MSE, indicating its superiority. As to the other models, they lose the modeling power due to their no account of dynamic cross-correlations. In particular, SFSL performs still better than the other two models thanks to its similar structure as DFSL. In contrast, though FPCAGra also considers sparsity for the cross-correlation structure, its correlation is based on the FPCA scores, which are yet extracted for every function separately. Since this feature extraction fails to consider the cross-correlations of different functions, it losses some information and consequently undermines the estimation. As to MFPCA, it does not take sparsity into account at all, and hence has the poorest performance.

We further compare the four models with sparsity structure in terms of the false subspace identification rate. In particular, for SFSL, for the estimated $\mathbf{b}_j (j = 1, \ldots, p)$ in every replication, we test if any function not belonging to the subspace of $\mathbf{Y}_{ij}$ has nonzero co-efficients. If so, we conclude that $\mathbf{Y}_{ij}$ is falsely identified. For DFSL, we use the average $\bar{\mathbf{b}}_j = \sum_{k=1}^{n} \mathbf{b}_j(t_k)/n$ for test, following the same procedure as SFSL. As to the two graphic models KerGra and FPCAGra, we test whether two functions from different subspaces have an edge. If so, considering the graph is undirected, we treat both of these two functions as falsely identified. Then we calculate the false identification rate for different models using 100 simulation replications. As Figure 10b shows, similar to the results of MSE, DFSL has the smallest false identification rate for different noise magnitude. Though KerGra follows DFSL closely for smaller noise magnitude, as the noise increases, their difference becomes larger. This indicates that with the same estimation MSE achieved, the price of KerGra is bigger than DFSL, with a large over-fitting. This over-fitting becomes severer for large noise magnitude, even rendering KerGra worse performance than SFSL. In addition, it is very interesting to see that in contrast to the other three models, the false identification rate of FPCAGra becomes smaller as the noise magnitude increases. This is because that FPCAGra prefers a sparser graph when the dependency structure of different functions become weaker, which is exactly the case with increased independent noise.

Finally, we test the cross-correlation change detection capability of DFSL and KerGra. In particular, for DFSL, we set the threshold as $c_{j0} = 3 \times \text{std}_j(c_{jk})$ and identify the change points $T_j$ of $\mathbf{Y}_j$. Then we calculate $C_k(k = 1, \ldots, n)$ and identify the system-level change points as those $t_k$ with $C_k \geq 1$. For KerGra, we use the similar identification procedure by simply changing $b_{jr}(t_k)$ in (14) as the smoothed partial cross-correlation of $\mathbf{Y}_j$ and $\mathbf{Y}_r$ at time $t_k$. In this way, $c_{j0}$ are expected to filter out some noisy dynamics caused by the Kernel smoothing. For the identified change points of DFSL and KerGra, only the change identifications occur near to the true change point, i.e., in the set $\{\tau_{s-1}, \tau_s, \tau_{s+1}\}(s = 1, \ldots, S-1)$ are accurate change point detections, while the other change point identifications are regarded as false detections. We report the average false change point detection with 100 simulation replications in Figure 10c. It shows that DFSL almost has no false change point detection for small noise magnitude. As the noise magnitude increases, the false detection increases, but is still satisfactorily small. Furthermore, DFSL has much smaller false change point detection than KerGra, especially for cases with large noise magnitude, indicating the robustness of DFSL. As to miss change point detection, since DFSL has constant zero mis-detection for all the simulation settings, we do not report the result here. However, KerGra does have nonzero miss detection, further demonstrating the superiority of DFSL.

## 5.3 Sensitivity Analysis

Finally, to have a complete understanding of DFSL, we evaluate its performance sensitivity to the number of functions per subspace. As Figure 11 shows, DFSL has smaller estimation MSE, false subspace identification rate and false change point detection rate overall, as the number of functions per subspace increases, indicating DFSL is especially beneficial for higher dimensional cases with larger number of functions. This is because as the number of functions per subspace increases, the point density per subspace increases, and the probability that one function can be well fitted by its subspace neighbors increases. Consequently, the fitting probability by functions from other subspaces decreases. This phenomenon is consistent with Theorem (3).

# 6 Case Studies

## 6.1 Motion Tracking

Now we reconsider the human gesture tracking experiments introduced in Section 1. In the experiment, this subject repeats the two gestures for nine times. Each of them is one sample $\mathbf{Y}_i(t_k) = [Y_{i1}(t_k), \ldots, Y_{i54}(t_k)]\,(k = 1, \ldots, n_i)$, where $n_i$ is the number of frames (i.e., the time length) of sample $i$. Since for different samples $n_i$ can be different, we first remove this non-synchronization among multiple samples using the dynamic time warping method (Keogh 2002) and make their time length equal to each other. Figure 14 plot one sample's fitting curves. They can capture the patterns of the 54 functions satisfactorily.

Based on the estimated $\hat{\mathbf{b}}_j(j = 1, \ldots, 54)$, we can calculate $C_k(1 \leq k \leq n)$ and estimate the potential change points of the cross-correlations of the 54 functions, as shown in Figure 15. By defining the system potential change points as those $\{k|C_k > 3 \times \mathrm{std}(C_k)\}$, $k = 111$, which is exactly the change point of this two gestures, is identified. Then we do subspace clustering for each of these two time segments separately. One thing to be noted is that if we cluster the 54 coordinates directly for this application, it is very possible to cluster the three coordinates of one joint into different clusters (subspaces), which is unreasonable to some degree. As such, here we propose to combine the three $\hat{\mathbf{b}}_j$ coming from one joint $J$ together as $\boldsymbol{\beta}_J = \sum_{j \in J} \hat{\mathbf{b}}_j / 3$, and use $\boldsymbol{\beta}_J(J = 1, \ldots, 18)$ to cluster the 18 joints. Since the number of clusters is hard to be pre-specified, here we apply the hierarchical clustering with the maximum within-cluster distance set to be 1.4. Specifically, we achieve six clusters for the first time segment and five clusters for the second one. The clustering results are visualized in Figure 13, where the joints from different clusters are denoted by different marker styles and colors. In particular, for the first segment, the six joints on the two arms are clustered together, since every three joints on each arm are connected together and the two arms move in the same way in this "bow up" gesture. Similarly, the six joints on the trunk are clustered together. As to the four joints on the kicks and feet, each of them are identified as one single subspace. This is because as shown in the video, the kicks have some "swing" movements, which yet do not appear in the movements of arms or trunks. As to the feet, they are not required to move as other joints, but only have some random fluctuation signals (as shown in Figure 12). Therefore, these two joints on the feet cannot be predicted well by other joints, and consequently are identified as two individual subspaces. As to the second segment, since this gesture requires the trunk, the right leg and the left arm to twist in the same way, these 12 joints are clustered together. The right arm is responsible for the "throw" action, so its joints are clustered together. As to the left leg, its two joints are identified as individual subspaces due to the same reason as the first segment.

In summary, the proposed DFSL can well describe the clusterwise cross-correlations of different joints (and coordinates), and can capture their dynamics effectively as well.

## 6.2    Manufacturing Process Monitoring

Now we consider another example from an advanced manufacturing system. In the system, seven sensors are used to monitor different process variables during the fabrication of every product sample. For simplification purpose, we denote these sensors as S1 to S7. For every sensor, it collects the signal of this process variable in every 0.1 second. Figure 16 illustrates their functional data for one product sample. It is clearly observed that some functions are quite similar with each other (such as S5, S6, and S7), demonstrating their strong correlations. Some other functions have quite diverse features (such as S1 and S5), indicating their weak cross-correlations. This illustrates that different sensors measure process variables from different sources and hence can be regarded as data coming from different subspaces sharing almost no similarity. Furthermore, during the fabrication, we usually need some on-off operations, such as filling materials into the reactor. These operations will artificially change the process variables, and consequently influence their correlation structure. As such, it is intuitive to apply DFSL to analyze this kind of systems.

In the dataset, we totally have $N = 46$ samples. Similar to the previous example, the profile length of different samples is different due to the fabrication inherent fluctuations. Therefore, we first remove the non-synchronization effect for different samples and set their time length as $n = 92$. Then we use DFSL to fit these 46 samples. The fitting result for one sample is shown in Figure 16. Unsurprisingly, DFSL can depict the patterns of these profiles with satisfactory accuracy. We further report the regression coefficients $\mathbf{b}_j(j = 1, \ldots, 7)$ in Figure 17. It shows that the coefficients are quite sparse, and most of them have a jump at around $k = 38$. Similar to 6.1, we calculate $C_k$ for every time point. As shown in Figure 18a, most of the identified change points are concentrated around $k = 38$. However, actually the true on-off operation occurs at $k = 35$. This delay is caused by the system operation delay itself. i.e., most process variables do not response to the operation until $k = 38$.

We further use the hierarchical clustering algorithm for sensor clustering. The affinity matrix is shown in Figure 18b, and the clustering result is quite consistent with engineering evaluations. The first four sensors belong to one subspace, and the other three belong to another subspace, demonstrating the efficiency of the proposed model once again.

# 7    Concluding remarks

Though multivariate functional data are common in many applications, they have not yet attracted enough attentions in the current research works. These functions are naturally high-dimensional data and have complex cross-correlations. For example, some functions share quite similar features (i.e., strong cross-correlations), while some others have quite diverse ones (i.e., very weak cross-correlations). Furthermore, their cross-correlation structure may change over time due to the system evolutions. Consequently, how to describe multivariate functions considering their complex and dynamic cross-correlations is a very challenging problem. With this regard, we propose a dynamic functional subspace learning method for multivariate functional data modeling. In particular, our model considers that different functions come from different subspaces. Only functions from the same subspace have nonzero cross-correlations with each other, while functions from different subspaces have no cross-correlations at all. Furthermore, we allow the subspace structure to change over time but regularize its change flexibility. Consequently we can describe the cross-correlation dynamics and also avoid over-parametrization. We also discuss the model inference in detail in terms of parameter estimation based on the fast iterative shrinkage-thresholding algorithm (FISTA), parameter tuning, and subspace recovering based on the smooth multi-channel functional PCA. Finally, some numerical studies together with two real case studies demonstrate the efficiency and applicability of the proposed methodology.

Along this research direction, there are several potential valuable extensions. Firstly, as shown in those two case studies, different functional samples may have different time length, or even irregular sampling time points. This is the common misalignment (deformation) problem. In these cases the current proposed model cannot be applied directly. Furthermore, this misalignment may introduce additional noise and functional dissimilarity into the data. To eliminate these problems, some transformations with data alignment techniques need to be incorporated into the model. Second, in this paper we temporarily assume $d_l < p_l$ for every subspace. When this assumption is violated for a certain subspace, it becomes unidentifiable. Then how to tackle this scenario deserves more research. Finally, how to use the proposed model to construct a statistical monitoring scheme to detect outlier functional samples is another future work direction.

# Appendices

In the Appendix, for case of presentation, we work on the case that $\mathbf{\Gamma}_j = \mathbf{I}$, while all of our methods and theoretical results can be extended to cases with general $\mathbf{\Gamma}$ follow the same procedures. For notation convenient, we redefine $\mathbf{b}_j(t_k) \in \mathcal{R}^{(p-1)\times 1}$ as $[b_{j1}(t_k), \ldots, b_{j(j-1)}(t_k), b_{j(j+1)}(t_k), \ldots, b_{jp}(t_k)]$, indicating the true cross-correlations at $t_k$. $\hat{\mathbf{b}}_j(t_k)$ is its correspondingly estimated one. Furthermore, we define $\boldsymbol{\beta}_j^s = \mathbf{b}_j(t_k), k \in [\tau_{s-1}, \tau_s)$ as true cross-correlations in the $s$ time segment.

## Appendix 1: Proof of Theorem 2

**Lemma 4.** *A matrix $\hat{\mathbf{b}}_j$ is optimal for the optimization of (8) if and only if there exist subgradient vectors $\hat{\mathbf{u}}_j(t_k) \in \partial||\hat{\mathbf{b}}_j(t_k) - \hat{\mathbf{b}}_j(t_{(k-1)})||_1$ and $\hat{\mathbf{v}}_j(t_k) \in \partial||\hat{\mathbf{b}}_j(t_k)||_1$ for $k = 1, \ldots, n$ that satisfy*

$$\sum_{i=1}^{N} \sum_{k=l}^{n} \mathbf{Y}_{i(-j)}(t_k)\langle \mathbf{Y}_{i(-j)}(t_k), \hat{\mathbf{b}}_j(t_k) - \mathbf{b}_j(t_k)\rangle - \sum_{i=1}^{N} \sum_{k=l}^{n} \mathbf{Y}_{i(-j)}(t_k)e_{ij}(t_k) + \lambda_1 \hat{\mathbf{u}}_j(t_k) + \lambda_2 \sum_{k=l}^{n} \hat{\mathbf{v}}_j(t_k) = 0,$$
(16)

for all $k = 1, \ldots, n$. Here $e_{ij}(t_k) = \sum_{r \neq j} \epsilon_{ir}(t_k) b_{jr}(t_k) - \epsilon_{ij}(t_k)$, and $\hat{\mathbf{u}}_j(t_1) = \hat{\mathbf{u}}_j(t_n) = 0$.

We follow the proof of Proposition 5 in Harchaoui and Lévy-Leduc (2010) and Theorem 2 in Kolar and Xing (2012). In particular, based on the union bound, we have

$$P[\max_{s \in 1, \ldots, S-1} |\tau_s - \hat{\tau}_s| > \delta_N] \leq \sum_{s=1}^{S-1} P[|\tau_s - \hat{\tau}_s| > \delta_N].$$

Then the theorem will hold up if we can prove that $P[|\tau_s - \hat{\tau}_s| > \delta_N] \to 0$ for all $s = 1, \ldots, S-1$. Define the set $A_{n,s}$ as

$$A_{N,s} = \{|\tau_s - \hat{\tau}_s| > \delta_N\},$$

and the set $C_N$ as

$$C_N = \{\max_{1 \leq s \leq S-1} |\tau_s - \hat{\tau}_s| < \frac{\Delta_{min}}{2}\},$$

where $\Delta_{min} = \min_{1 \leq s \leq S} |\tau_s - \tau_{s-1}|$. Then it is enough to prove that $P[A_{N,s} \cap C_N] \to 0$ and that $P[A_{N,s} \cap \bar{C}_c] \to 0$.

Let use first consider the proof of $P[A_{N,s} \cap C_N] \to 0$. Note that $C_N$ implies that $\tau_{s-1} \leq \hat{\tau}_s \leq \hat{\tau}_{s+1}$ for all $s = 1, \ldots, S-1$. We first assume $\hat{\tau}_s \leq \tau_s$, then use (16) twice with $k = \tau_s$ and $k = \hat{\tau}_s$ and apply the triangle inequality. We have

$$2\lambda_1 p \geq || \sum_{i=1}^{N} \sum_{k=\hat{\tau}_s}^{\tau_{s-1}} \mathbf{Y}_{i(-j)}(t_k) \langle \mathbf{Y}_{i(-j)}(t_k), \hat{\mathbf{b}}_j(t_k) - \mathbf{b}_j(t_k) \rangle - \sum_{i=1}^{N} \sum_{k=\hat{\tau}_k}^{\tau_k - 1} \mathbf{Y}_{i(-j)}(t_k) e_{ij}(t_k) ||_1. \quad (17)$$

Recall the true $\mathbf{b}_j(t_k), k \in [\tau_{s-1}, \tau_s)$ as $\boldsymbol{\beta}_j^s$, and define the estimated $\hat{\mathbf{b}}_j(t_k), k \in [\hat{\tau}_{s-1}, \hat{\tau}_s)$ as

$\hat{\boldsymbol{\beta}}_j^s$. This yields the event $C_{N,s}$ defined as follows, occurs with probability one:

$$C_{N,s} = \{2\lambda_1 p + (\tau_s - \hat{\tau}_s)\lambda_2 p \geq \|\sum_{i=1}^{N}\sum_{k=\hat{\tau}_s}^{\tau_s-1}\mathbf{Y}_{i(-j)}(t_k)\langle\mathbf{Y}_{i(-j)}(t_k), \boldsymbol{\beta}_j^s - \boldsymbol{\beta}_j^{s+1}\rangle\|_1 \tag{18}$$

$$- |\sum_{i=1}^{N}\sum_{k=\hat{\tau}_s}^{\tau_s-1}\mathbf{Y}_{i(-j)}(t_k)\langle\mathbf{Y}_{i(-j)}(t_k), \boldsymbol{\beta}_j^{s+1} - \hat{\boldsymbol{\beta}}_j^{s+1}\rangle\|_1$$

$$- \|\sum_{i=1}^{N}\sum_{k=\hat{\tau}_k}^{\tau_k-1}\mathbf{Y}_{i(-j)}(t_k)e_{ij}(t_k)\|_1\} \tag{19}$$

$$=: \|R_1\|_1 - \|R_2\|_1 - \|R_3\|_1.$$

Using that $P[A_{N,s} \cap C_N] = P[A_{N,s} \cap C_N \cap C_{N,s}]$, we can get

$$P[A_{N,s} \cap C_N] \leq P[A_{N,s} \cap C_N \cap \{2\lambda_1 p + (\tau_s - \hat{\tau}_s)p\lambda_2 \geq \frac{1}{3}\|R_1\|_1\}] \tag{20}$$

$$+ P[A_{N,s} \cap C_N \cap \{\|R_2\|_1 \geq \frac{1}{3}\|R_1\|_1\}]$$

$$+ P[A_{N,s} \cap C_N \cap \{\|R_3\|_1 \geq \frac{1}{3}\|R_1\|_1\}]$$

$$=: P[A_{N,s,1}] + P[A_{N,s,2}] + P[A_{N,s,3}].$$

According to Lemma 6, we can upper bound $P[A_{N,s,1}]$ with

$$P[2p\lambda_1 + (\tau_s - \hat{\tau}_s)\sqrt{p}\lambda_2 \geq N\frac{\phi_{s0}}{27}(\tau_s - \hat{\tau}_s)\xi_{min}] + 2\exp(-\delta_N N + 2\log N + 2\log n).$$

With the assumption $(N\delta_N p\xi_{min})^{-1}\lambda_1 \to 0$ and $(N\sqrt{p}\xi_{min})^{-1}\lambda_2 \to 0$ as $N \to \infty$, we have that $P[A_{N,s,1}] \to 0$ as $N \to \infty$.

To show $P[A_{N,s,2}]$ converges to zero, define $\bar{\tau}_s = \lfloor 2^{-1}(\tau_s + \tau_{s+1})\rfloor$. With $C_n$, we have $\hat{\tau}_{s+1} > \bar{\tau}_s$. Consequently, we have $\mathbf{b}_j(t_k) = \hat{\boldsymbol{\beta}}_j^{s+1}$ for $k \in [\tau_s, \bar{\tau}_s]$. Using (4) with $k = \tau_s$ and $k = \bar{\tau}_s$, we have

$$2p\lambda_1 + (\bar{\tau}_s - \tau_s)\sqrt{p}\lambda_2 \geq \|\sum_{i=1}^{N}\sum_{k=\tau_s}^{\bar{\tau}_s-1}\mathbf{Y}_{i(-j)}(t_k)\langle\mathbf{Y}_{i(-j)}(t_k), \boldsymbol{\beta}_j^{s+1} - \hat{\boldsymbol{\beta}}_j^{s+1}\rangle\|_1 - \|\sum_{i=1}^{N}\sum_{k=\tau_s}^{\bar{\tau}_s-1}\mathbf{Y}_{i(-j)}(t_k)e_{ij}(t_k)\|_1.$$

With Lemma 6 on the display above, we have

$$||\boldsymbol{\beta}_j^{s+1} - \hat{\boldsymbol{\beta}}_j^{s+1}||_1 \le \sqrt{p}\frac{36p\lambda_1 + 18(\bar{\tau}_s - \tau_s)\sqrt{p}\lambda_2 + 18||\sum_{i=1}^{N}\sum_{k=\tau_s}^{\bar{\tau}_s-1}\mathbf{Y}_{i(-j)}(t_k)e_{ij}(t_k)||_1}{(\tau_{s+1} - \tau_s)\phi_{s0}N}, \quad (21)$$

which holds with probability at least $1 - 2\exp(-\Delta_{min}N/4 + 2\log N + 2\log n)$.

Furthermore, based on Lemma 6, we have $N\phi_{s0}(\tau_s - \hat{\tau}_s)\xi_{min}/9 \le \sqrt{p}||R_1||_1$ and $||R_2||_1 \le N(\tau_s - \hat{\tau}_s)9\phi_{s1}\sqrt{p}||\boldsymbol{\beta}_j^{s+1} - \hat{\boldsymbol{\beta}}_j^{s+1}||_1$ with probability at least $1 - 4\exp(-N\delta_N/2 + 2\log N + 2\log n)$. Consequently, with (21), $P[A_{n,s,2}]$ is upper bounded by

$$P[\lambda_1 \ge c_1 p^{-2}\phi_{s0}^2\phi_{s1}^{-1}\Delta_{min}N\xi_{min}] + P[\lambda_2 \ge c_2 p^{-1.5}\phi_{s0}^2\phi_{s1}^{-1}\xi_{min}N] \quad (22)$$

$$+ P[(\bar{\tau}_j^{s+1} - \tau_j^s)^{-1}N^{-1}||\sum_{i=1}^{N}\sum_{k=\tau_s}^{\bar{\tau}_s-1}\mathbf{Y}_{i(-j)}(t_k)e_{ij}(t_k)||_1 \ge c_3\xi_{min}p^{-1}\phi_{s0}^2\phi_{s1}^{-1}]$$

$$+ c_4\exp(-N\delta_n/2 + 2\log n + 2\log N).$$

The first, second and the last term converge to zero as $N \to \infty$. For the third term, it converges to zero with the rate $\exp(-c_6\log N)$ since $\log N/N \to 0$.

Now we show $P[A_{N,s,3}]$ converges to zero. Since $N\phi_{s0}(\tau_s - \hat{\tau}_s)\xi_{min}/9 \le \sqrt{p}||R_1||_1$ with probability $1 - 2\exp(-N\delta_N/2 + 2\log n + 2\log N)$, we have the upper bound of $P[A_{N,s,3}]$ as

$$P[\frac{\phi_{s0}\xi_{min}}{27} \le \sqrt{p}\frac{||\sum_{i=1}^{N}\sum_{k=\hat{\tau}_k}^{\tau_k-1}\mathbf{Y}_{i(-j)}(t_k)e_{ij}(t_k)||_1}{N(\tau_s - \hat{\tau}_s)}] + 2\exp(-N\delta_N/2 + 2\log n + 2\log N),$$

which, according to Lemma 7, converges to zero as $\log N/N \to 0$. As to the case with $\hat{\tau}_s > \tau_s$, we can show the above proof in a similar way. Consequently, we have $P[A_{N,s} \cap C_N] \to 0$ as $N \to \infty$.

We proceed to show that $P[A_{N,s} \cap \bar{C}_N] \to 0$ as $N \to \infty$. Recall $\bar{C}_N = \{\max_s |\hat{\tau}_s - \tau_s| \ge \Delta_{min}/2\}$, We now split it into three events,

$$D_N^l = \{\exists s, \hat{\tau}_s \le \tau_{s-1}\} \cap \bar{C}_N,$$

$$D_N^m = \{\forall s, \tau_{s-1} < \hat{\tau}_s < \tau_{s+1}\} \cap \bar{C}_N,$$

$$D_N^r = \{\exists s, \hat{\tau}_s \ge \tau_{s+1}\} \cap \bar{C}_N.$$

Then we have $P[A_{N,s} \cap \bar{C}_N] = P[A_{N,s} \cap D_N^l] + P[A_{N,s} \cap D_N^m] + P[A_{N,s} \cap D_N^r]$.

We first focus on $P[A_{N,s} \cap D_N^m]$ and consider the case where $\hat{\tau}_s \leq \tau_s$, since the case with $\tau_s \leq \hat{\tau}_s$ can be addressed in a similar way. Note that

$$
\begin{aligned}
P[A_{N,s} \cap D_N^m] \leq & P[A_{N,s} \cap \{(\hat{\tau}_{s+1} - \tau_s) \geq \frac{\Delta_{min}}{2}\} \cap D_n^m] \\
& + \sum_{g=s+1}^{S-1} P[\{(\tau_g - \hat{\tau}_g) \geq \frac{\Delta_{min}}{2}\} \cap \{(\hat{\tau}_{g+1} - \tau_g) \geq \frac{\Delta_{min}}{2}\} \cap D_N^m].
\end{aligned}
\tag{23}
$$

We first bound the first term in (23). Using (16) with $k = \hat{\tau}_s$ and $k = \tau_s$, we have

$$
\frac{||\boldsymbol{\beta}_j^s - \hat{\boldsymbol{\beta}}_j^{s+1}||_1}{p} \leq \frac{18p\lambda_1 + 9(\tau_s - \hat{\tau}_{s+1})\sqrt{p}\lambda_2 + 9||\sum_{i=1}^N \sum_{k=\hat{\tau}_k}^{\tau_k - 1} \mathbf{Y}_{i(-j)}(t_k)e_{ij}(t_k)||_1}{N\phi_{s0}(\tau_s - \hat{\tau}_s)},
\tag{24}
$$

with probability at least $1 - 2\exp(-\delta_N N/2 + 2\log N + 2\log n)$ Define $\bar{\tau}_s = \lfloor(\tau_s + \tau_{s+1})/2\rfloor$. Using (16) with $k = \bar{\tau}_s$ and $k = \tau_s$, we have

$$
\begin{aligned}
\frac{||\boldsymbol{\beta}_j^s - \boldsymbol{\beta}_j^{s+1}||_1}{p} \leq & \frac{18p\lambda_1 + 9(\bar{\tau}_s - \tau_s)\sqrt{p}\lambda_2 + 9||\sum_{i=1}^N \sum_{k=\tau_k}^{\bar{\tau}_k - 1} \mathbf{Y}_{i(-j)}(t_k)e_{ij}(t_k)||_1}{N\phi_{s0}(\bar{\tau}_s - \tau_s)} \\
& + 81\phi_{s0}^{-1}\phi_{s1}||\boldsymbol{\beta}_j^s - \hat{\boldsymbol{\beta}}_j^{s+1}||_1,
\end{aligned}
\tag{25}
$$

with probability at least $1 - c_1\exp(-\delta_N N + 2\log n + 2\log N)$. Combining the two above inequalities and using Lemma 7, we can upper bound the first term in (23) with

$$
P[\xi_{min}\phi_{s0}N\delta_N \leq c_1 p^2\lambda_1] + P[\phi_{s0}N \leq c_2 p^{1.5}\lambda_2] + P[\xi_{min}\sqrt{N\delta_N} \leq c_3 p^2\sqrt{\log N}] + c_4\exp(-c_5\log N).
$$

Under the conditions of the theorem, all the terms converge to zero. Using the similar way, we can prove the other items in (23) converge to zero. Finally, we can conclude $P[A_{N,s} \cap D_N^m] \to 0$ as $N \to \infty$. As to $P[A_{N,s} \cap D_N^l]$, it is upper bounded by

$$
P[D_N^l] \leq \sum_{s=1}^S 2^{s-1} P[\{\max_{l=1,\ldots,S} \hat{\tau}_l \leq \tau_{l-1}\} = s]
\tag{26}
$$

$$
\leq 2^{S-1} \sum_{s=1}^{S-1} \sum_{l\leq s} P[\{\tau_l - \hat{\tau}_l \geq \frac{\Delta_{min}}{2}\} \cap \{\hat{\tau}_{l+1} - \tau_l \geq \frac{\Delta_{min}}{2}\}].
\tag{27}
$$

With the same arguments as those used to bound (23), we can prove $P[D_N^l] \to 0$ and $P[D_N^r] \to 0$ as $N \to \infty$. Consequently, we can show $P[A_{N,s} \cap \bar{C}_N] \to 0$.

## Appendix 2: Proof of Theorem 3

With Theorem 2 satisfied, we are working on the event

$$\mathcal{E} = \{ \max_{s=1,\dots,S} |\hat{\tau}_s - \tau_s| \leq \delta_N \}.$$

Define $\mathcal{B}^s = [\tau_{s-1}, \tau_s)$ and the corresponding estimated $\hat{\mathcal{B}}^s = [\hat{\tau}_{s-1}, \hat{\tau}_s)$, then for $k \in \hat{\mathcal{B}}^s$, we have

$$Y_{ij}(t_k) = \mathbf{Y}_{i(-j)}(t_k)\boldsymbol{\beta}_j^s + w_{ij}(t_k) + e_{ij}(t_k),$$

where $w_{ij}(t_k) = \mathbf{Y}_{i(-j)}(t_k)(\mathbf{b}_j(t_k) - \boldsymbol{\beta}_j^s)$ is the bias. For $k \in \mathcal{B}^s \cap \hat{\mathcal{B}}^s$, the bias $w_{ij}(t_k) = 0$, otherwise the bias is distributed with zero mean and bounded variance under the Assumption (A6). Since $\hat{\boldsymbol{\beta}}_j^s$ is an optimal solution of (8), it satisfies

$$\sum_{i=1}^N \mathbf{Y}_{i(-j)}^{\hat{\mathcal{B}}^s\prime} \mathbf{Y}_{i(-j)}^{\hat{\mathcal{B}}^s}(\hat{\boldsymbol{\beta}}_j^s - \boldsymbol{\beta}_j^s) - \sum_{i=1}^N \mathbf{Y}_{i(-j)}^{\hat{\mathcal{B}}^s\prime}(\mathbf{w}_{ij}^{\hat{\mathcal{B}}^s} + \mathbf{e}_{ij}^{\hat{\mathcal{B}}^s}) + \lambda_1(\hat{\mathbf{u}}_j(t_{\hat{\tau}_{s-1}}) - \hat{\mathbf{u}}_j(t_{\hat{\tau}_s})) + \lambda_2|\hat{\mathcal{B}}^s|\hat{\mathbf{v}}_j(t_{\hat{\tau}_{s-1}}) = 0,$$
(28)

where $\mathbf{Y}_{i(-j)}^{\hat{\mathcal{B}}^s} \in \mathcal{R}^{|\hat{\mathcal{B}}^s| \times (p-1)}$ are the observations in the $s^{th}$ estimated segment $\hat{\mathcal{B}}^s$. $\mathbf{w}_{ij}^{\hat{\mathcal{B}}^s}, \mathbf{e}_{ij}^{\hat{\mathcal{B}}^s} \in \mathcal{R}^{|\hat{\mathcal{B}}^s| \times 1}$ are the stacked $w_{ij}(t_k)$ and $e_{ij}(t_k)$ in $\hat{\mathcal{B}}^s$ respectively.

**Lemma 5.** *Suppose $\hat{\mathbf{b}}_j$ is a solution of (8), with the associated segment points $\hat{T} = \{\hat{\tau}_s, s = 1, \dots, S-1\}$. Suppose that the subgradient vectors satisfy $|\hat{v}_{jr}(t_k)| < 1$ for all $r \notin S(\hat{\mathbf{b}}_j(t_k))$ where $S(\hat{\mathbf{b}}_j(t_k))$ denotes the set of nonzero elements of $\hat{\mathbf{b}}_j(t_k)$. Then any other solution $\tilde{\mathbf{b}}_j$ with the same time segment points as $\hat{T}$ satisfies $\tilde{b}_{jr}(t_k) = 0$ for $r \notin \mathcal{S}(\hat{\mathbf{b}}_j(t_k))$.*

According to Lemma 5, we consider designing vectors $\check{\boldsymbol{\beta}}_j^s, \check{\mathbf{u}}_j(t_{\hat{\tau}_s})$ and $\check{\mathbf{v}}_j(t_{\hat{\tau}_s})$ with the associated $\check{T}$ that satisfy (28) and $\check{\beta}_{jr}^s = 0$ for $r \notin S(\boldsymbol{\beta}_j^s)$. Then if we can verify the subdifferential vectors, $\check{\mathbf{u}}_j(t_{\hat{\tau}_s})$ and $\check{\mathbf{v}}_j(t_{\hat{\tau}_s})$ are dual feasible, we can prove any other solution $\tilde{\boldsymbol{\beta}}_j^s$ with the same time segment points as $\check{T}$ satisfies $\tilde{\beta}_{jr}^s = 0$ for $r \notin \mathcal{S}(\boldsymbol{\beta}_j^s)$.

Denote $M_j^s = \mathcal{S}(\boldsymbol{\beta}_j^s)$ as the set of functions that belong to the same subspace as $\mathbf{Y}_{ij}^{\mathcal{B}^s}$, and $N_j^s$ as the set of functions that do not belong to $M_j^s$. Now we consider the following restricted optimization problem,

$$\min_{\substack{\boldsymbol{\beta}_j^s, s=1,...,S \\ \tilde{\boldsymbol{\beta}}_{j,N_j^s}^s=0}} \sum_{s=1}^{S} ||\mathbf{Y}_{ij}^{\hat{\mathcal{B}}^s} - \mathbf{Y}_{i(-j)}^{\hat{\mathcal{B}}^s}\boldsymbol{\beta}_j^s||_2^2 + 2\lambda_1 \sum_{s=2}^{S} ||\boldsymbol{\beta}_j^s - \boldsymbol{\beta}_j^{s-1}||_1 + 2\lambda_2 \sum_{s=1}^{S} |\hat{\mathcal{B}}^s|||\boldsymbol{\beta}_j^s||_1, \tag{29}$$

where the vector $\boldsymbol{\beta}_{j,N_j^s}^s$ is constrained to be $\mathbf{0}$. Let $\{\check{\boldsymbol{\beta}}_j^s\}$ be a solution to the restricted optimization problem (29). Its subgradient vectors are $\check{\mathbf{u}}_j(t_{\hat{\tau}_{s-1}}) \in \partial||\check{\boldsymbol{\beta}}_j^s - \check{\boldsymbol{\beta}}_j^{s-1}||_1$, $\check{\mathbf{u}}_j(t_{\hat{\tau}_s}) \in \partial||\check{\boldsymbol{\beta}}_j^{s+1} - \check{\boldsymbol{\beta}}_j^s||_1$, and $\check{\mathbf{v}}_j(t_{\hat{\tau}_{s-1}}) = \text{sign}(\check{\boldsymbol{\beta}}_j^s)$.

It is obvious that the vectors $\check{\boldsymbol{\beta}}_j^s$, $\check{\mathbf{u}}_j(t_{\hat{\tau}_s})$ and $\check{\mathbf{v}}_j(t_{\hat{\tau}_s})$ satisfy (28). Furthermore, $\check{\mathbf{u}}_j(t_{\hat{\tau}_{s-1}})$ and $\check{\mathbf{u}}_j(t_{\hat{\tau}_{s-1}})$ are elements of the subdifferential and hence dual feasible. To show $\check{\boldsymbol{\beta}}_j^s$ is also a solution to (8), we need to show that $\check{\mathbf{v}}_j(t_{\hat{\tau}_{s-1}})$ is also dual feasible, i.e., $||\check{\mathbf{v}}_{j,N_j^s}(t_{\hat{\tau}_{s-1}})||_\infty < 1$. Then any other solution $\tilde{\boldsymbol{\beta}}_j^s$ to (8) will satisfy $\tilde{\boldsymbol{\beta}}_{j,N_j^s}^s = 0$.

Form (28), we obtain an explicit formula of $\boldsymbol{\beta}_{j,M_j^s}^s$,

$$\check{\boldsymbol{\beta}}_{j,M_j^s}^s = \boldsymbol{\beta}_{j,M_j^s}^s + \left(\sum_{i=1}^{N} \mathbf{Y}_{i(-j),M_j^s}^{\check{\mathcal{B}}^s\prime}\mathbf{Y}_{i(-j),M_j^s}^{\check{\mathcal{B}}^s}\right)^{-1} \left(\sum_{i=1}^{N} \mathbf{Y}_{i(-j),M_j^s}^{\check{\mathcal{B}}^s\prime}(\mathbf{w}_{ij,M_j^s}^{\check{\mathcal{B}}^s} + \mathbf{e}_{ij,M_j^s}^{\check{\mathcal{B}}^s})\right.$$
$$\left. - \lambda_1(\hat{\mathbf{u}}_{j,M_j^s}(t_{\hat{\tau}_{s-1}}) - \hat{\mathbf{u}}_{j,M_j^s}(t_{\hat{\tau}_s})) - \lambda_2|\check{\mathcal{B}}^s|\hat{\mathbf{v}}_{j,M_j^s}(t_{\hat{\tau}_{s-1}})\right).$$

Then plug the above equation into (28), we have that $||\check{\mathbf{v}}_{j,N_j^s}(t_{\hat{\tau}_{s-1}})||_\infty < 1$, if $\max_{r \in N_j^s}|Y_r| < 1$ where

$$Y_r = \sum_{i=1}^{N} \mathbf{Y}_{ir}^{\check{\mathcal{B}}^s\prime}\mathbf{Y}_{i(-j),M_j^s}^{\check{\mathcal{B}}^s} \left(\sum_{i=1}^{N} \mathbf{Y}_{i(-j),M_j^s}^{\check{\mathcal{B}}^s\prime}\mathbf{Y}_{i(-j),M_j^s}^{\check{\mathcal{B}}^s}\right)^{-1} \left(\check{\mathbf{v}}_{j,M_j^s}(t_{\hat{\tau}_{s-1}}) + \frac{\lambda_1(\hat{\mathbf{u}}_{j,M_j^s}(t_{\hat{\tau}_{s-1}}) - \hat{\mathbf{u}}_{j,M_j^s}(t_{\hat{\tau}_s}))}{\lambda_2|\hat{\mathcal{B}}^s|}\right.$$
$$\tag{30}$$
$$\left. + \sum_{i=1}^{N} \mathbf{Y}_{i(-j),M_j^s}^{\check{\mathcal{B}}^s}\frac{\mathbf{w}_{ij}^{\check{\mathcal{B}}^s} + \mathbf{e}_{ij}^{\check{\mathcal{B}}^s}}{\lambda_2|\hat{\mathcal{B}}^s|}\right) + \sum_{i=1}^{N} \mathbf{Y}_{ir}^{\check{\mathcal{B}}^s\prime}\frac{\mathbf{w}_{ij}^{\check{\mathcal{B}}^s} + \mathbf{e}_{ij}^{\check{\mathcal{B}}^s}}{\lambda_2|\hat{\mathcal{B}}^s|}.$$

As $N \to \infty$, we have $\sum_{i=1}^{N} \mathbf{Y}_{i(-j),M_j^s}^{\check{\mathcal{B}}^s\prime}\mathbf{Y}_{i(-j),M_j^s}^{\check{\mathcal{B}}^s} \to N\frac{|\hat{\mathcal{B}}^s|}{n}(1+\sigma^2)\mathbf{I}$, and hence its inverse goes

to $n(N|\hat{\mathcal{B}}^s|)^{-1}(1+\sigma^2)^{-1}\mathbf{I}$. As such, the first term of Equation (30) has the upper bound

$$\frac{n}{N|\hat{\mathcal{B}}^s|(1+\sigma^2)}\sum_{i=1}^{N}||\mathbf{Y}_{ir}^{\hat{\mathcal{B}}^s\prime}\mathbf{Y}_{i(-j),M_j^s}^{\hat{\mathcal{B}}^s}||_\infty \left( ||\check{\mathbf{v}}_{j,M_j^s}(t_{\hat{\tau}_{s-1}})||_\infty + ||\frac{\lambda_1(\hat{\mathbf{u}}_{j,M_j^s}(t_{\hat{\tau}_{s-1}}) - \hat{\mathbf{u}}_{j,M_j^s}(t_{\hat{\tau}_s}))}{\lambda_2|\hat{\mathcal{B}}^s|}||_\infty \right.$$

$$\left. +||\sum_{i=1}^{N}\mathbf{Y}_{i(-j),M_j^s}^{\hat{\mathcal{B}}^s}\frac{\mathbf{w}_{ij}^{\hat{\mathcal{B}}^s}+\mathbf{e}_{ij}^{\hat{\mathcal{B}}^s}}{\lambda_2|\hat{\mathcal{B}}^s|}||_\infty \right).$$

We have

$$||\mathbf{Y}_{ir}^{\hat{\mathcal{B}}^s\prime}\mathbf{Y}_{i(-j),M_j^s}^{\hat{\mathcal{B}}^s}||_\infty \leq \sum_{g:\hat{\mathcal{B}}^s\cap\mathcal{B}^g\neq\emptyset}||\mathbf{Y}_{ir}^{\hat{\mathcal{B}}^s\cap\mathcal{B}^g\prime}\mathbf{Y}_{i(-j),M_j^s}^{\hat{\mathcal{B}}^s\cap\mathcal{B}^g}||_\infty$$

$$\leq \sum_{g:\hat{\mathcal{B}}^s\cap\mathcal{B}^g\neq\emptyset}||\mathbf{X}_{ir}^{\hat{\mathcal{B}}^s\cap\mathcal{B}^g\prime}\mathbf{X}_{i(-j),M_j^s}^{\hat{\mathcal{B}}^s\cap\mathcal{B}^g}||_\infty + \sum_{g:\hat{\mathcal{B}}^s\cap\mathcal{B}^g\neq\emptyset}||\mathbf{X}_{ir}^{\hat{\mathcal{B}}^s\cap\mathcal{B}^g\prime}\boldsymbol{\epsilon}_{i(-j),M_j^s}^{\hat{\mathcal{B}}^s\cap\mathcal{B}^g}||_\infty$$

$$+ \sum_{g:\hat{\mathcal{B}}^s\cap\mathcal{B}^g\neq\emptyset}||\boldsymbol{\epsilon}_{ir}^{\hat{\mathcal{B}}^s\cap\mathcal{B}^g\prime}\mathbf{X}_{i(-j),M_j^s}^{\hat{\mathcal{B}}^s\cap\mathcal{B}^g}||_\infty + \sum_{g:\hat{\mathcal{B}}^s\cap\mathcal{B}^g\neq\emptyset}||\boldsymbol{\epsilon}_{ir}^{\hat{\mathcal{B}}^s\cap\mathcal{B}^g\prime}\boldsymbol{\epsilon}_{i(-j),M_j^s}^{\hat{\mathcal{B}}^s\cap\mathcal{B}^g}||_\infty.$$

According to Lemma 8 and 9, with probability at least $1 - \frac{c_1}{(M_j^s)^2}$, we have the following inequalities,

$$||\mathbf{X}_{ir}^{\hat{\mathcal{B}}^s\cap\mathcal{B}^g\prime}\mathbf{X}_{i(-j),M_j^s}^{\hat{\mathcal{B}}^s\cap\mathcal{B}^g}||_\infty \leq \frac{|\hat{\mathcal{B}}^s\cap\mathcal{B}^g|}{n}\sqrt{24}\log p \max_{r,l=1,\dots,L^g}\frac{\text{aff}(\mathcal{S}_r^g,\mathcal{S}_l^g)}{\sqrt{\max(d_l^g,d_r^g)}},$$

$$||\mathbf{X}_{ir}^{\hat{\mathcal{B}}^s\cap\mathcal{B}^g\prime}\boldsymbol{\epsilon}_{i(-j),M_j^s}^{\hat{\mathcal{B}}^s\cap\mathcal{B}^g}||_\infty \leq \frac{|\hat{\mathcal{B}}^s\cap\mathcal{B}^g|}{n}2\sigma\sqrt{\frac{2\log|\hat{\mathcal{B}}^s\cap\mathcal{B}^g|}{|\hat{\mathcal{B}}^s\cap\mathcal{B}^g|}},$$

$$||\boldsymbol{\epsilon}_{ir}^{\hat{\mathcal{B}}^s\cap\mathcal{B}^g\prime}\mathbf{X}_{i(-j),M_j^s}^{\hat{\mathcal{B}}^s\cap\mathcal{B}^g}||_\infty \leq \sqrt{\frac{|\hat{\mathcal{B}}^s\cap\mathcal{B}^g|}{n}}\sqrt{24}\frac{\log p}{\sqrt{p}}||\boldsymbol{\epsilon}_{ir}^{\hat{\mathcal{B}}^s\cap\mathcal{B}^g}||_2,$$

$$||\boldsymbol{\epsilon}_{ir}^{\hat{\mathcal{B}}^s\cap\mathcal{B}^g\prime}\boldsymbol{\epsilon}_{i(-j),M_j^s}^{\hat{\mathcal{B}}^s\cap\mathcal{B}^g}||_\infty \leq 2\sigma\sqrt{\frac{2\log|\hat{\mathcal{B}}^s\cap\mathcal{B}^g|}{n}}||\boldsymbol{\epsilon}_{ir}^{\hat{\mathcal{B}}^s\cap\mathcal{B}^g}||_2.$$

According to Lemma 10, we have

$$P[||\boldsymbol{\epsilon}_{ir}^{\hat{\mathcal{B}}^s\cap\mathcal{B}^g}||_2 \geq \sqrt{\frac{(1+\omega)\sigma^2}{n}|\hat{\mathcal{B}}^s\cap\mathcal{B}^g|}] \leq \exp(-\frac{1}{8}|\hat{\mathcal{B}}^s\cap\mathcal{B}^g|\omega^2).$$

As such, we have

$$\frac{n}{N|\hat{\mathcal{B}}^s|(1+\sigma^2)}\sum_{i=1}^{N}||\mathbf{Y}_{ir}^{\hat{\mathcal{B}}^s\prime}\mathbf{Y}_{i(-j),M_j^s}^{\hat{\mathcal{B}}^s}||_\infty \leq \frac{1}{1+\sigma^2}\left(\sqrt{24}\log p\max_{\substack{r,l=1,\ldots,L^g\\g:\hat{\mathcal{B}}^g\cap\mathcal{B}^g\neq\emptyset}}\frac{\text{aff}(\mathcal{S}_r^g,\mathcal{S}_l^g)}{\sqrt{\max d_r^g,d_l^g}}+2c_0\sigma\right.$$
$$\left.+\sqrt{24}\frac{\log p}{\sqrt{p}}\sqrt{1+\omega}\sigma+2c_0\sigma^2\sqrt{1+\omega}\right),$$

with probability $1-c_1/(M_j^s)^2-c_2\exp(-\Delta_{min}\omega^2/8)$, under the event $\mathcal{E}$. (Here $c_0$ is smaller than 1). Furthermore, we have

$$||\check{\mathbf{v}}_{j,M_j^s}(t_{\hat{\tau}_{s-1}})||_\infty \leq 1,$$
$$||\frac{\lambda_1(\hat{\mathbf{u}}_{j,M_j^s}(t_{\hat{\tau}_{s-1}})-\hat{\mathbf{u}}_{j,M_j^s}(t_{\hat{\tau}_s}))}{\lambda_2|\hat{\mathcal{B}}^s|}||_\infty \leq \frac{2\lambda_1}{\lambda_2(\hat{\tau}_s-\hat{\tau}_{s-1})}. \tag{31}$$

As long as $\lambda_1/\lambda_2 \to 0$ as $N \to \infty$, we can bound (31) to 0. At last, according to Lemma 7, we have with probability at least $1-c_3\exp(-c_4N)$,

$$||\sum_{i=1}^{N}\mathbf{Y}_{i(-j),M_j^s}^{\hat{\mathcal{B}}^s}\frac{\mathbf{w}_{ij}^{\hat{\mathcal{B}}^s}+\mathbf{e}_{ij}^{\hat{\mathcal{B}}^s}}{\lambda_2|\hat{\mathcal{B}}^s|}||_2 \leq \frac{\phi_{s1}^{1/2}\sqrt{1+C}\sigma_{w+e}}{\lambda_2}\sqrt{|M_j^s|(1+\log N)N|\hat{\mathcal{B}}^s|},$$

where $\sigma_{w+e}^2$ is the variance of $w_{ij}(t_k)+e_{ij}(t_k)$. As $\lambda_2\sqrt{N\log N}\to\infty$, we have the above item bounded to 0. Similarly, we can bound the second item in (30) to 0. Consequently, we have that as long as there is a positive constant $\omega$ that satisfies

$$\sqrt{1+\omega} \leq \frac{1+\sigma^2-\sqrt{24}\log p(\max_{\substack{r,l=1,\ldots,L^g\\g=1,\ldots,S}}\frac{\text{aff}(\mathcal{S}_r^g,\mathcal{S}_l^g)}{\sqrt{\max(d_r^g,d_l^g)}})-2\sigma}{\sqrt{24}\frac{\log p}{\sqrt{p}}\sigma+2\sigma^2}, \tag{32}$$

we have with probability at least $1-c_1/(M_j^s)^2-c_2\exp(-\Delta_{min}\omega^2/8)$, $\max_{r\in N_j^s}|Y_r|<1$ and $\check{\boldsymbol{\beta}}_{j,N_j^s}=0$. Consequently, we can prove any solution $\tilde{\boldsymbol{\beta}}_j^s$ with the same segment points $\check{T}$ satisfies $\tilde{\beta}_{jr}^s=0$ for $r\notin\mathcal{S}(\boldsymbol{\beta}_j^s)$.

## Appendix 3: Other Useful Results

Here we show some additional results that are useful for the derivations in Appendix 1 and Appendix 2.

Denote the functional data in the $s^{\text{th}}$ time segment as $\mathbf{Y}_i^s = \left[ \mathbf{Y}_{i(1)}^s, \ldots, \mathbf{Y}_{i(L^s)}^s \right]$ with the $l^{\text{th}}$ subspace data $\mathbf{Y}_{i(l)}^s \in \mathcal{R}^{n_s \times p_l^s}$. Denote $\boldsymbol{\Phi}^s = \left[ \boldsymbol{\Phi}_1^s, \ldots, \boldsymbol{\Phi}_p^s \right]$, $\mathbf{A}_i^s = \text{diag}(\mathbf{A}_{i(1)}^s, \ldots, \mathbf{A}_{i(L^s)}^s)$ and $\mathbf{E}_i^s = \left[ \epsilon_{i1}^s, \ldots, \epsilon_{ip}^s \right]$. Then we generate

$$\mathbf{Y}_i^s = \mathbf{X}_i^s + \mathbf{E}_i^s,$$

$$\mathbf{X}_i^s = \boldsymbol{\Phi}^s \mathbf{A}_i^s \quad l = 1, \ldots, L^s,$$

for $k = \tau_{s-1}+1, \ldots, \tau_s$. $\mathbf{E}_{i(l)}^s = \left[ \boldsymbol{\epsilon}_{i1}^s, \ldots, \boldsymbol{\epsilon}_{ip_i^s}^s \right]$ is the noise function, with $\boldsymbol{\epsilon}_{ij}^s \sim N_{n_s}(\mathbf{0}, \sigma^2/n\mathbf{I})(j = 1, \ldots, p)$. Denote

$$\mathbf{K}^s = \text{E}(\mathbf{Y}_i^{s'} \mathbf{Y}_i^s) = \text{E}(\mathbf{A}_i^{s'} \boldsymbol{\Phi}^{s'} \boldsymbol{\Phi}_k^s \mathbf{A}_i^s) + \frac{n_s}{n} \sigma^2 \mathbf{I}. \tag{33}$$

When every column of $\mathbf{A}_{il}^s$ is sampled at random from the unit sphere of $\mathcal{R}^{d_l \times 1}$, we have $\mathbf{K}^s = \frac{n_s}{n}(1 + \sigma^2)\mathbf{I}$. Define $\phi_{s1} = \Lambda_{max}(\mathbf{K}^s)$ and $\phi_{s0} = \Lambda_{min}(\mathbf{K}^s)$. Let $\hat{\mathbf{K}}^s = N^{-1}(m - l + 1)^{-1} \sum_{i=1}^N \sum_{k=l}^m \mathbf{Y}_i(t_k)' \mathbf{Y}_i(t_k)$ be the empirical estimation of $\mathbf{K}^s$ with $\tau_{s-1} < l < m < \tau_s$. There are two crude bounds for the eigenvalues of $\hat{\mathbf{K}}^s$

$$P(\Lambda_{\max}(\hat{\mathbf{K}}^s) \geq 9\phi_{s1}) \leq 2\exp(-N(m - l + 1)/2),$$

$$P(\Lambda_{\min}(\hat{\mathbf{K}}^s) \leq \phi_{s0}/9) \leq 2\exp(-N(m - l + 1)/2).$$

**Lemma 6.** *For any* $v_n N > p$, *we have*

$$P\left[ \max_{\substack{\tau_{s-1} < l < m \leq \tau_s \\ m - l > v_n}} \Lambda_{max}(\frac{\sum_{i=1}^N \sum_{k=l}^m \mathbf{Y}_i^{s'}(t_k)\mathbf{Y}_i^s(t_k)}{N(m - l + 1)}) \geq 9\phi_{s1} \right] \leq \exp(-v_n N/2 + 2\log(\tau_s - \tau_{s-1}) + 2\log(N)),$$

and

$$P\left[\max_{\substack{\tau_{s-1}<l<m\le\tau_s \\ m-l>v_n}}\Lambda_{min}(\frac{\sum_{i=1}^N\sum_{k=l}^m \mathbf{Y}_i^{s'}(t_k)\mathbf{Y}_i^s(t_k)}{N(m-l+1)})\le\phi_{s0}/9\right]\le\exp(-v_nN/2+2\log(\tau_s-\tau_{s-1})+2\log(N)).$$

**Lemma 7.** *Recall* $e_{ij}(t_k)=\sum_{r\ne j}\epsilon_{ir}(t_k)b_{jr}(t_k)-\epsilon_{ij}(t_k)$, *we know* $e_{ij}(t_k)$ *is also normal distributed with mean* 0 *and bounded variance, denoted as* $\gamma^2$. *Then if* $v\ge C\log N$ *for some constant* $C>16$,

$$P\left[\bigcap_{s=1,\dots,S}\bigcap_{\substack{\tau_{s-1}<l<m<\tau_s \\ m-l>v}}\left\{\frac{1}{N(m-l+1)}||\sum_{i=1}^N\sum_{k=l}^m\mathbf{Y}_i(t_k)e_{ij}(t_k)||_1\le\frac{p\gamma\phi_1^{1/2}\sqrt{1+C}}{\sqrt{N(m-l+1)}}\sqrt{1+C\log N}\right\}\right]$$

(34)

$$\ge 1-c_1\exp(-c_2\log N),$$

*for some constants* $c_1,c_2>0$. *Here* $\phi_1=\max_s\phi_{s1}$.

# Appendix 4: Standard Inequalities in Probability

**Lemma 8.** *Suppose the unit-norm vector* $\mathbf{x}_{ij}\in\mathcal{R}^{n\times 1}$ *is drawn uniformed at random from* $\mathcal{S}_r$, *and* $\mathbf{X}_{i(l)}\in\mathcal{R}^{n\times p_l}$ *are* $p_l$ *unit-norm vectors drawn uniformly at random from* $\mathcal{S}_l$, *then we have*

$$||\mathbf{X}_{i(l)}'\mathbf{x}_{ij}||_\infty\le\sqrt{\log a\log b}\frac{\sqrt{d_l\wedge d_r}}{\sqrt{d_ld_r}}aff(\mathcal{S}_l,\mathcal{S}_r),$$

(35)

*with probability at least* $1-\frac{2}{\sqrt{a}}-\frac{2p_l}{\sqrt{b}}$.

**Lemma 9.** *Suppose* $\mathbf{Z}\in\mathcal{R}^{n\times p}$ *has iid* $N(0,1)$ *entries and let* $\mathbf{x}\in\mathcal{R}^{n\times 1}$ *a unit-norm vector. Then*

$$||\mathbf{Z}'\mathbf{x}||_\infty\le 2\sqrt{2\log n},$$

*with probability at least* $1-2/p^2$.

**Lemma 10.** *For a $\chi_n^2$ distribution with $n$ degrees of freedom, it obeys*

$$P[\chi_n^2 \geq (1 + \omega)n] \leq \exp(-n\omega^2/8).$$

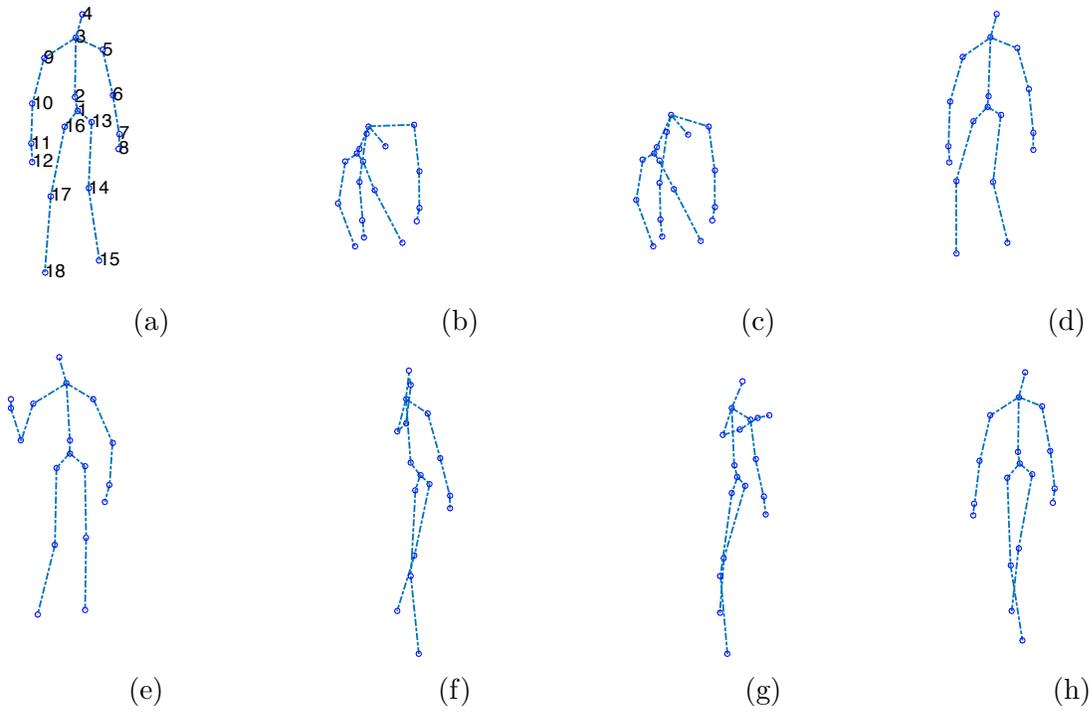

Figure 1: Snapshots of the joints at time point (a) $t_{25}$, (b) $t_{50}$, (c) $t_{75}$, (d) $t_{100}$, (e) $t_{125}$, (f) $t_{150}$, (g) $t_{175}$, (h) $t_{200}$. Specifically, the first four time points (a to d) belong to the first gesture "bow up", and the second four time points (e to h) belong to the second gesture "throw".

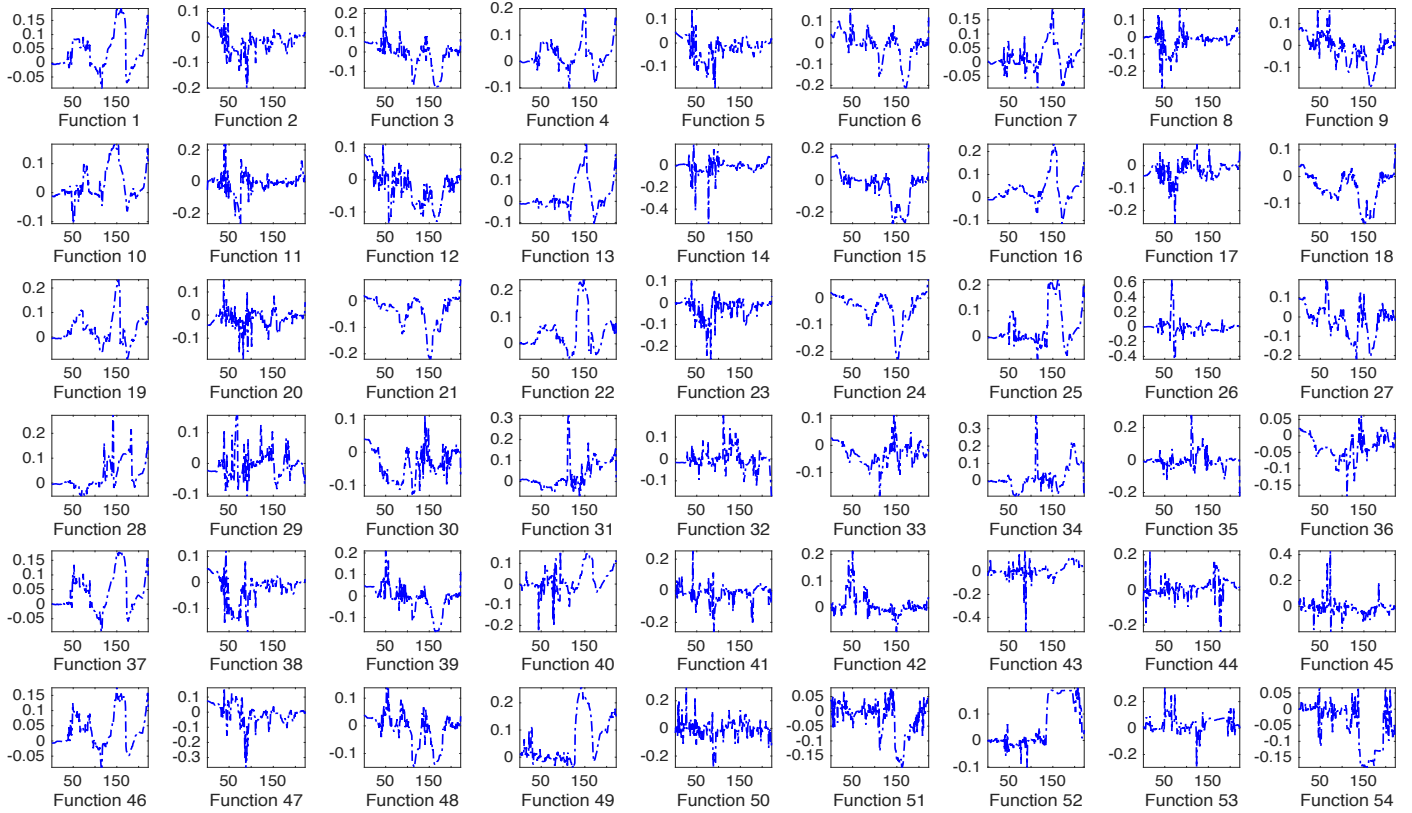

Figure 2: Profiles of the 54 coordinates for one sample. In particular, Functions $\{3j-2, 3j-1, 3j\}$ above correspond to the functions of the $\{x, y, z\}$ coordinates of joint $j$, for $j = 1, \ldots, 18$.

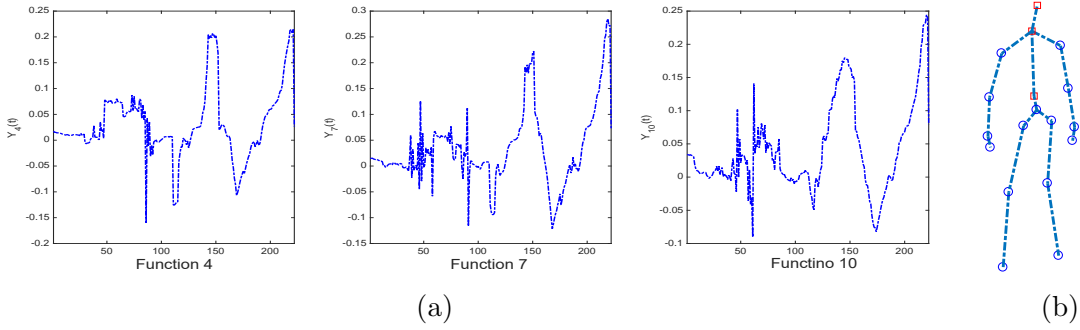

(a)                                                                    (b)

Figure 3: (a) Profiles of Functions 4, 7, and 10. They show similar patterns with each other, demonstrating their strong cross-correlations. (b) The corresponding joints of Functions 4, 7, and 10 (denoted in red).

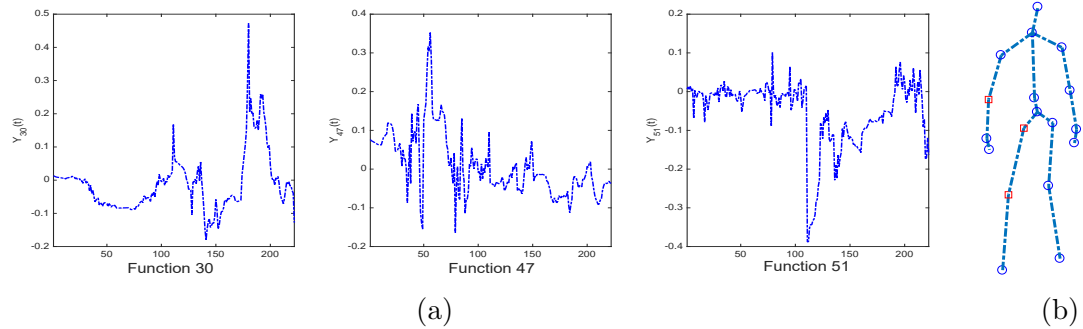

(a)                                                                    (b)

Figure 4: (a) Profiles of Functions 30, 47, and 51. They show quite different patterns with each other, demonstrating their weak cross-correlations. (b) The corresponding joints of Functions 30, 47, and 51 (denoted in red).

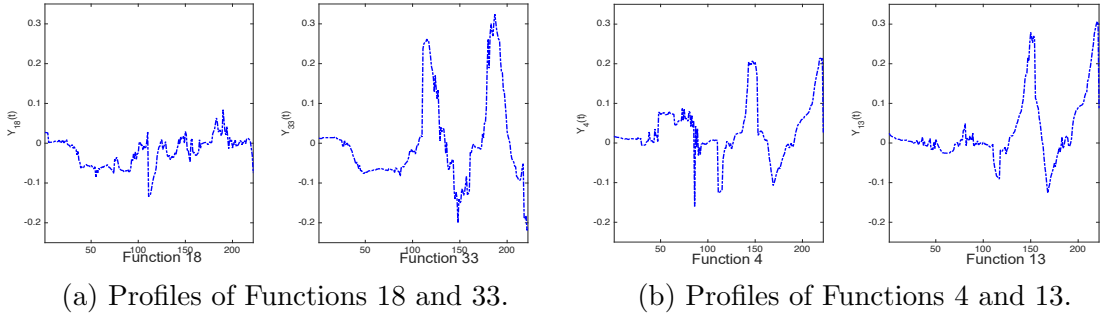

(a) Profiles of Functions 18 and 33.

(b) Profiles of Functions 4 and 13.

Figure 5: (a) Functions 18 and 33 have similar patterns for the first 110 time points; (b) Functions 4 and 13 and similar patterns for the later 138 time points.

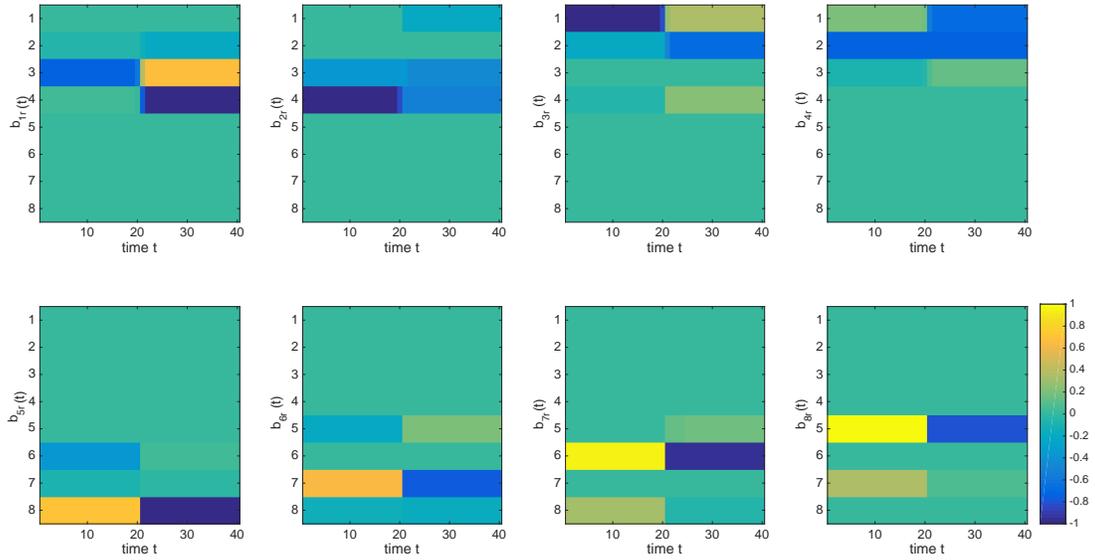

Figure 6: Estimated $\mathbf{b}_j(j = 1, \ldots, p)$ for Model (I).

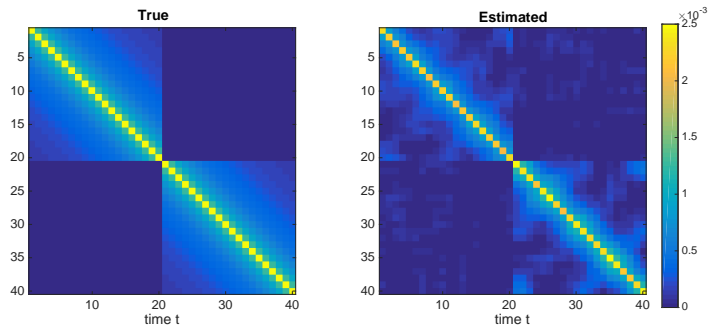

Figure 7: Estimated $\boldsymbol{\Sigma}_1$ of Model (I).

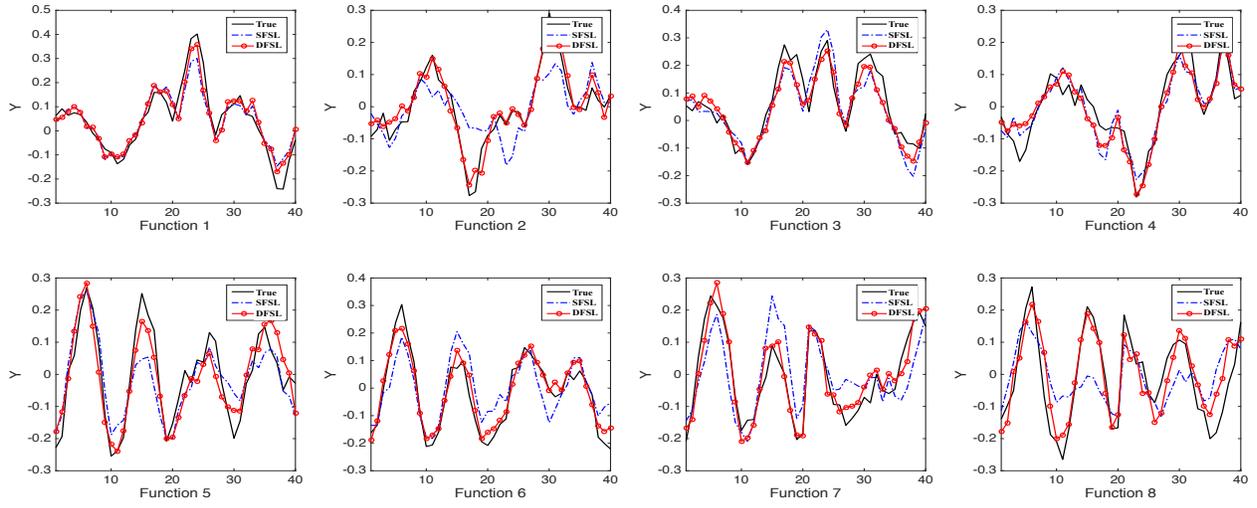

(a) Model (I)

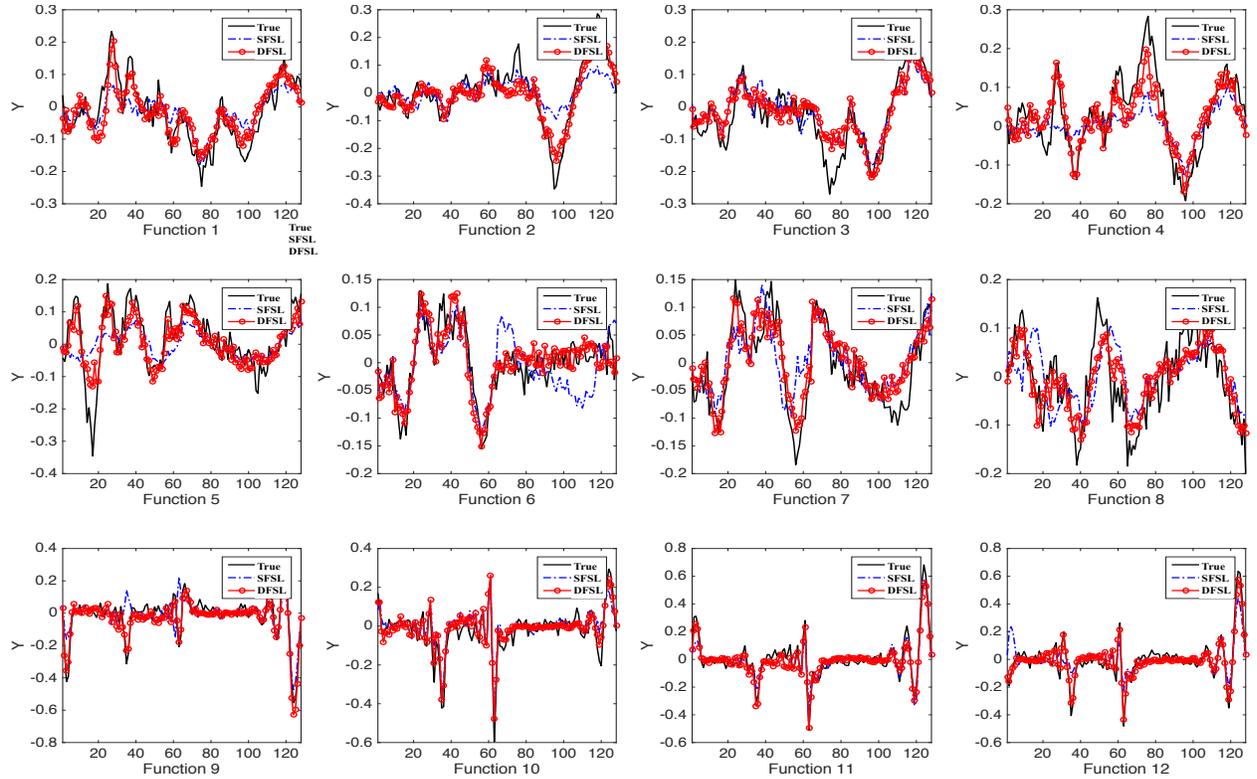

(b) Model (II)

Figure 8: Estimation results of DFSL and SFSL for Model (I) and Model (II).

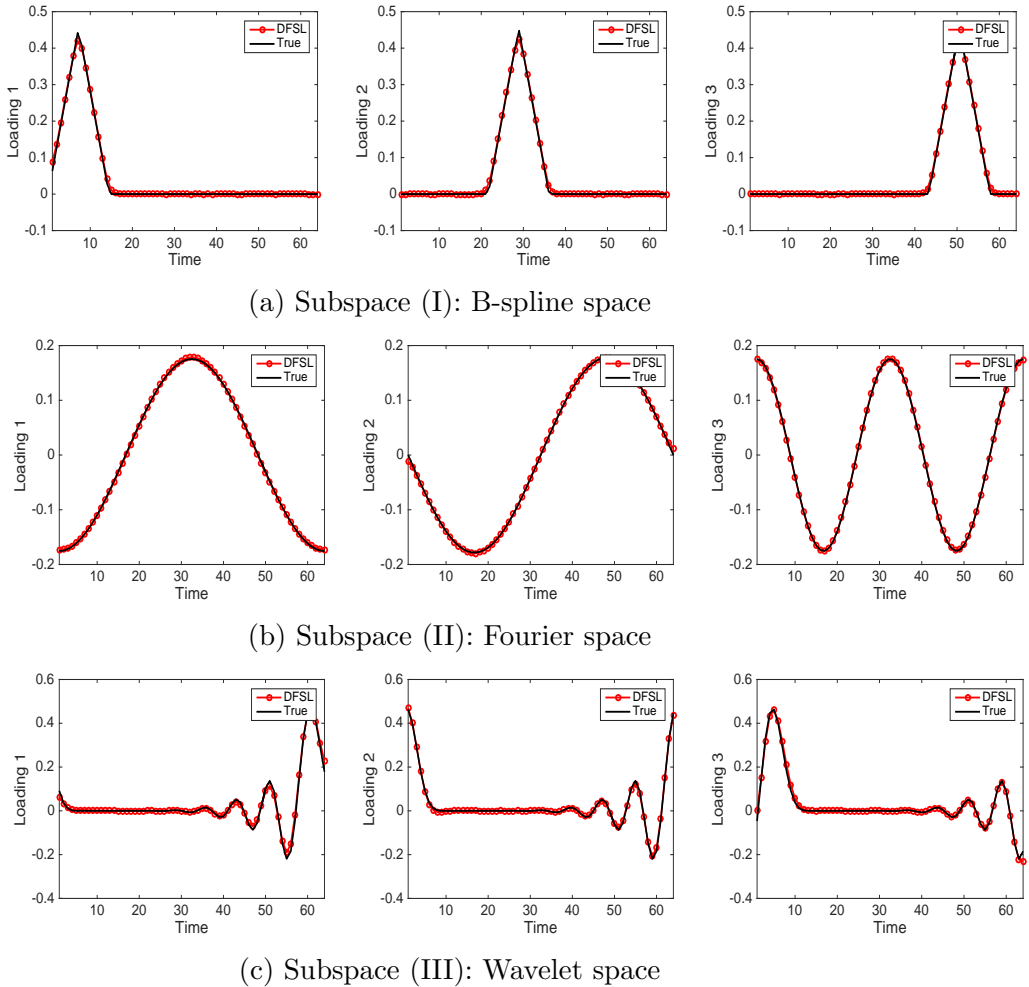

(a) Subspace (I): B-spline space

(b) Subspace (II): Fourier space

(c) Subspace (III): Wavelet space

Figure 9: Extracted basis functions of the three subspaces for Model (II) segment 3.

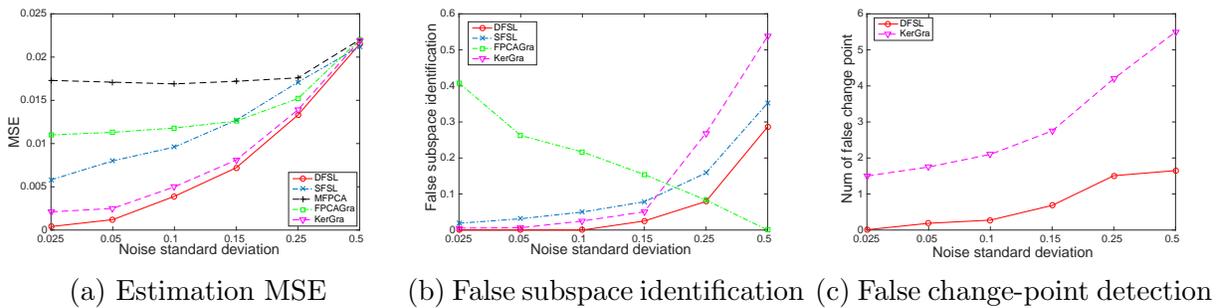

(a) Estimation MSE     (b) False subspace identification   (c) False change-point detection

Figure 10: Modeling results for Model (I).

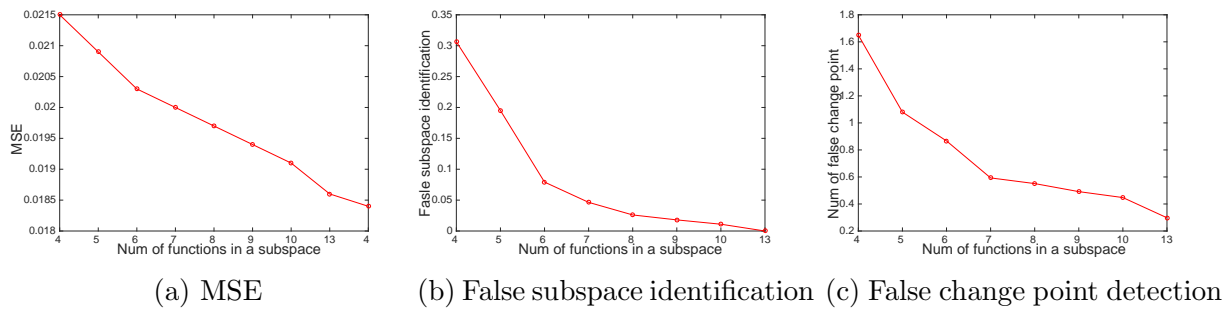

(a) MSE  (b) False subspace identification  (c) False change point detection

Figure 11: Modeling results of DFSL as the number of functions per subspace increases.

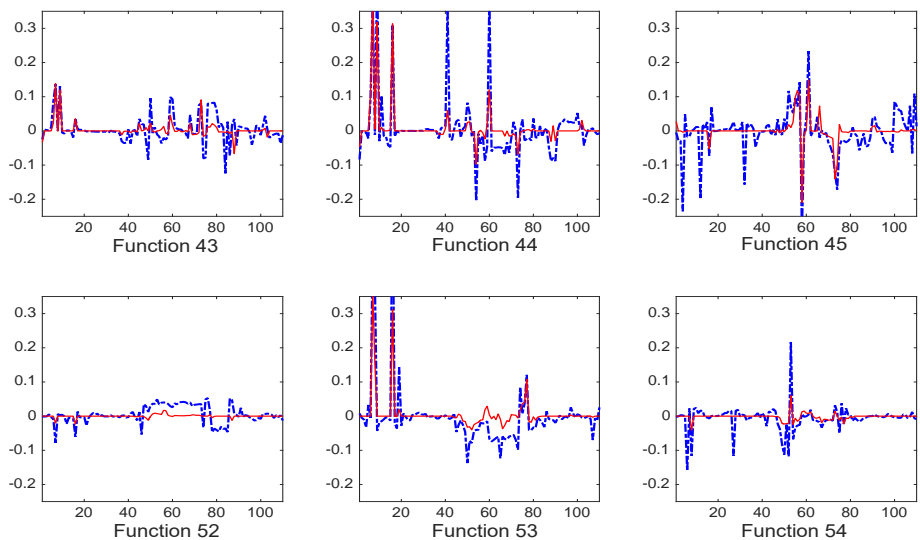

Figure 12: The profiles of the three coordinates of the left foot (the first row), and the profiles of three coordinates of the right foot (the second row) for the first segment $[1, 110]$.

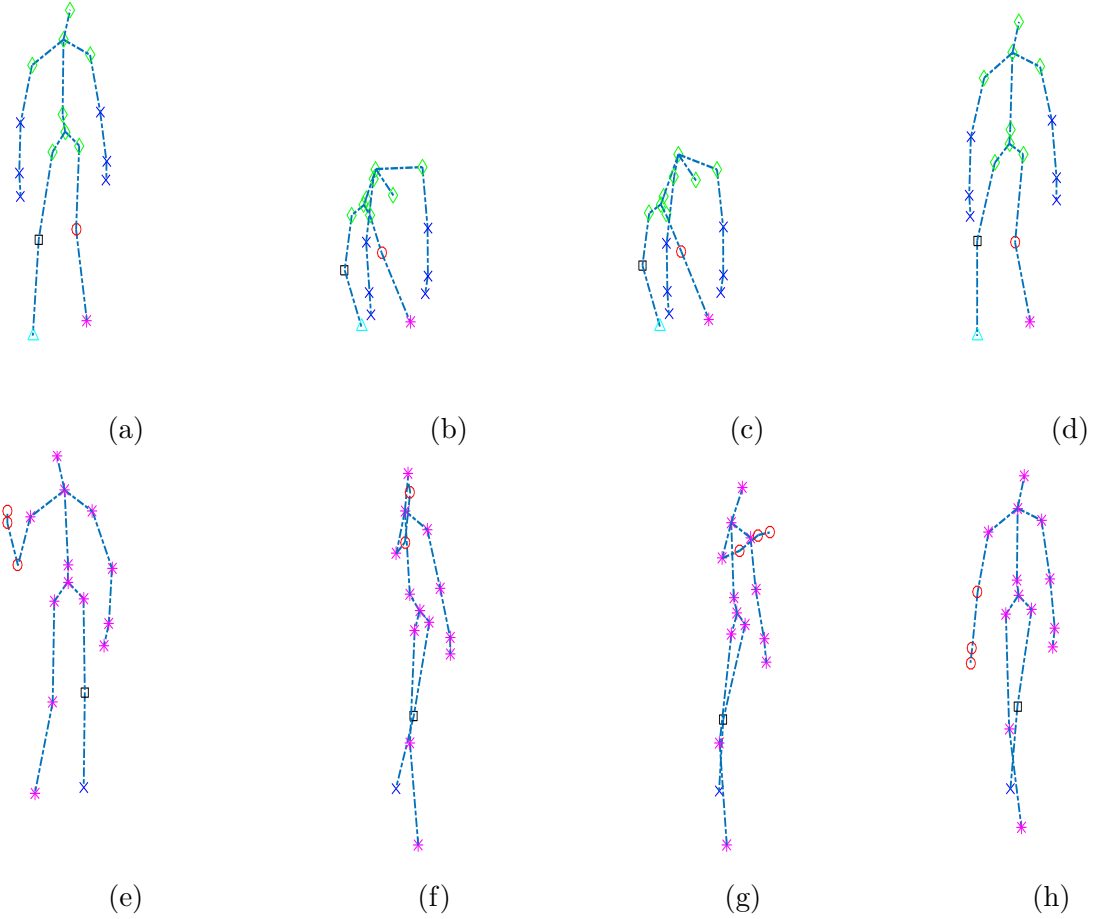

Figure 13: The joint clustering results for the two motions with the same snapshots as Figure 1. In particular, the joints of the same cluster are drawn in the same color and marker.

Figure 14: Profiles of the 54 coordinates for one sample. The blue dashed lines are true curves, and the red solid lines are estimated ones based on the proposed dynamic functional subspace learning (DFSL).

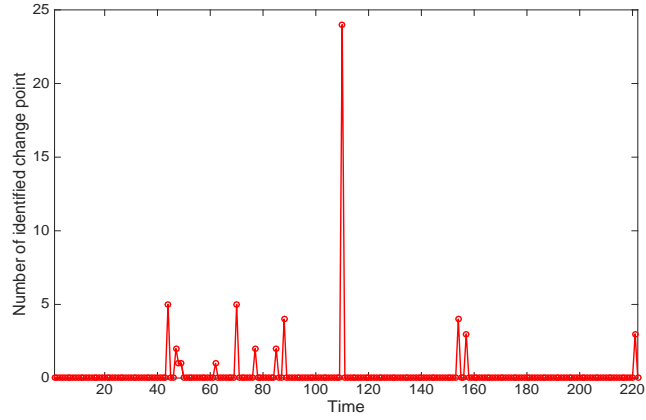

Figure 15: Number of Identified change points.

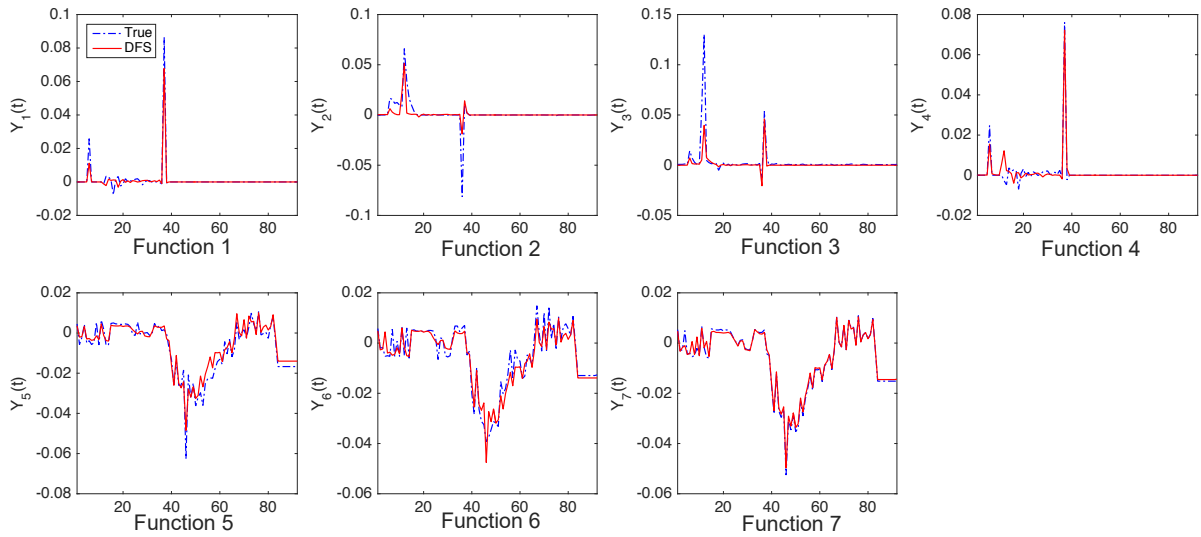

Figure 16: Profiles of the seven sensors in an advanced manufacturing system.

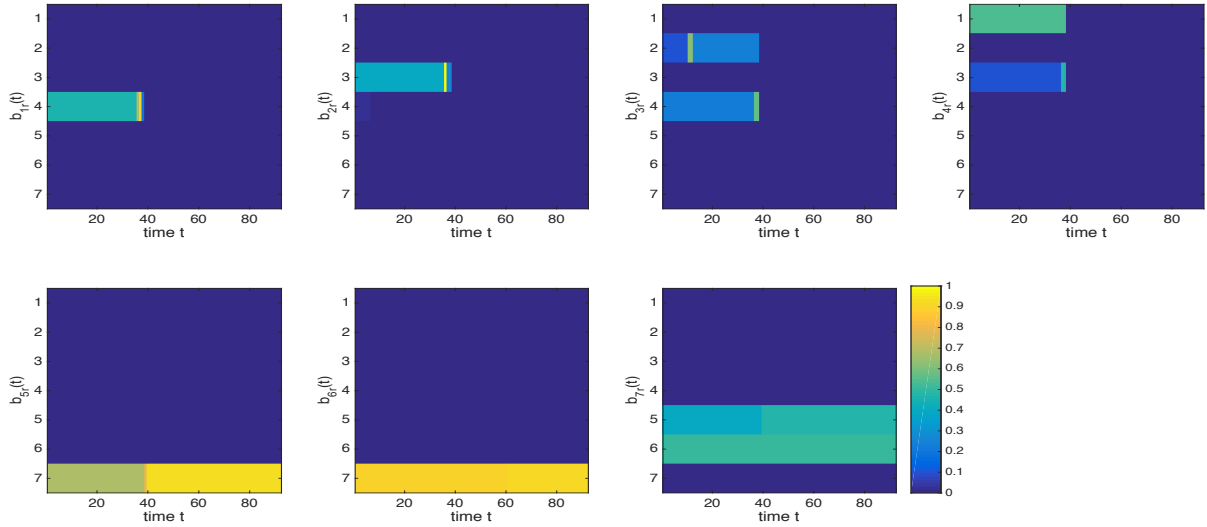

Figure 17: The estimated $\hat{\mathbf{b}}_j (j = 1, \ldots, 7)$ for the 7 sensors.

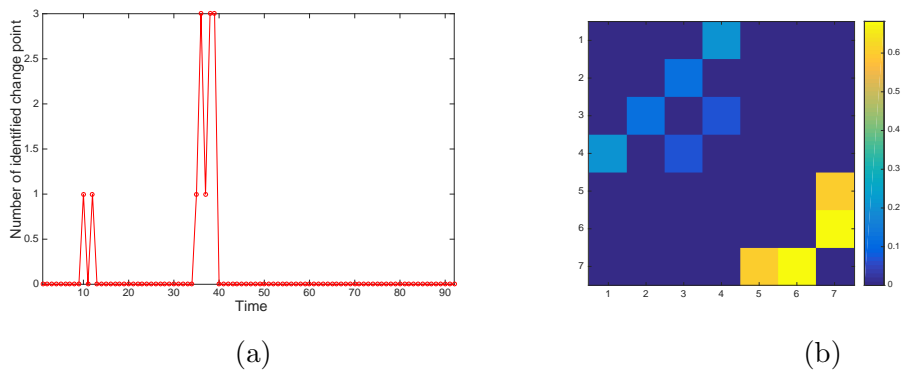

(a)                                                    (b)

Figure 18: (a) The number of identified change points for every time point; (b) The affinity matrix for the manufacturing system data.